\documentclass{article}
\usepackage{iclr2027_conference,times}

\usepackage{amsmath,amsfonts,bm}

\def\eqref#1{equation~\ref{#1}}

\def\1{\bm{1}}

\DeclareMathAlphabet{\mathsfit}{\encodingdefault}{\sfdefault}{m}{sl}
\SetMathAlphabet{\mathsfit}{bold}{\encodingdefault}{\sfdefault}{bx}{n}

\usepackage{hyperref}
\hypersetup{
  hidelinks,
  pdftitle={Keep the Effect, Drop the Actor: Programmable Effect-to-Execution World-Action Models},
  pdfauthor={Junyi Hu, Zhewen He, Zhenhua Li, Yi Fang}
}
\usepackage{url}
\usepackage{booktabs}
\usepackage{multirow}
\usepackage{graphicx}
\usepackage{xcolor}
\usepackage{amsmath,amssymb}
\usepackage{subcaption}
\usepackage{wrapfig}
\usepackage{placeins}
\usepackage{algorithm}
\usepackage{algpseudocode}

\newcommand{\Estar}{E^{\star}}
\newcommand{\Gstar}{G^{\star}}
\newcommand{\method}{PEWAM}
\newcommand{\best}[1]{\textbf{#1}}

\title{Keep the Effect, Drop the Actor:\\Programmable Effect-to-Execution World-Action Models}

\author{%
Junyi Hu\textsuperscript{1,*}\quad
Zhewen He\textsuperscript{1}\quad
Zhenhua Li\textsuperscript{1,2}\quad
Yi Fang\textsuperscript{1,2,*}\\[4pt]
{\normalfont\textsuperscript{1}New York University Abu Dhabi}\\
{\normalfont\textsuperscript{2}Chatsign Technology}\\[3pt]
{\normalfont\small\textsuperscript{*}Corresponding authors: Junyi Hu and Yi Fang.}\\
{\normalfont\small\href{mailto:junyihu095@gmail.com}{\texttt{junyihu095@gmail.com}},
\href{mailto:yf23@nyu.edu}{\texttt{yf23@nyu.edu}}}
}

\iclrfinalcopy

\begin{document}
\maketitle
\lhead{}
\renewcommand{\headrulewidth}{0pt}

\begin{abstract}
A robot demonstration records two things in the same frames: what happened to the objects, and how one
particular arm made it happen. We condition on the first. A demonstration is compiled into an \emph{effect
program}: the 3D keypoint trajectories of the objects that moved, two points marking where each was held,
and the configuration the scene ends in, with the demonstrator removed. \method{}, a 71.5M-parameter
world-action model, generates effect, robot execution, action and terminal state as four streams with
independent flow-matching times, so clamping a program and sampling the execution turns inference into
programming, re-solved closed loop from the live scene. On held-out LIBERO-Goal tasks, one demonstration's
program completes 40 of 90 episodes, where the same backbone given a goal image or language, and published
demonstration-conditioned methods, complete at most 19; on three of Meta-World's held-out classes it exceeds
the best published results, though not on the five-class mean. Because a program is a set of coordinates, a person can edit it: the placement
follows a shifted terminal state and the grasp turns with rotated contact points. The same program runs on
four robot arms without retraining, and after a push, re-solving completes 23 of 60 episodes where replaying
the demonstration completes 6. On a Franka arm, fine-tuned on real demonstrations of other tasks, programs compiled from single human videos complete 36 of 40 trials, against 22 for the same backbone conditioned on the video's last frame as a goal image.
\end{abstract}

\section{Introduction}
\label{sec:intro}
Showing a robot what to do is often easier than describing it, and a single demonstration is a direct way to
specify a new manipulation task~\citep{instantpolicy,kat,osviwm,vivla}. A demonstration, however,
records two things in the same frames: what happened to the objects, and how one particular arm made it
happen. Watch someone put a bowl on a stove and you learn where the bowl should end up, roughly how it
travels and where it can be held; you do not learn to bend your elbow the way they bent theirs. Most systems
that learn from demonstrations keep the two together. Policies that transfer the demonstrator's motion keep
the body and treat the world as context~\citep{demodiffusion,atm,pointpolicy}, and video-action models
prompted with a demonstration predict both at once, as one pixel-level future~\citep{hydra0,zerowam}. The
task is then entangled with one body's motion: no variable states it on its own, so it can be neither edited nor given to another robot.

This viewpoint exposes a limitation of both dominant formulations. Conditioning on the demonstrator's
trajectory or video~\citep{uwm,atm,instantpolicy} ties the task to one execution, which is valid only for the
arm, the grasp and the layout that produced it. Methods built on object motion do separate the object from
the actor~\citep{im2flow2act,generalflow,track2act,dream2flow}, but they generate that motion from language
or video and execute it with a planner, a controller or a per-task policy. What is missing is a task interface
that leaves the actor out, that one multi-task model accepts with no parameter update, and that a person can edit.

\begin{figure}[t]
\centering
\includegraphics[width=\textwidth]{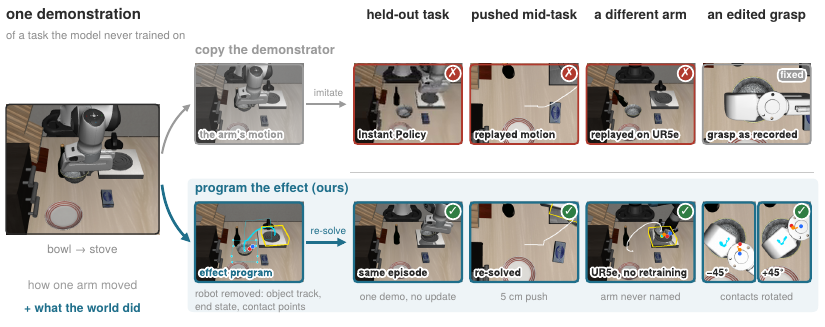}
\caption{\textbf{Keep the effect, drop the actor.} A demonstration records how one arm moved and what happened
to the world. Copying the demonstrator (top) ties the task to that motion: a published imitator fails the held-out task, replayed actions miss the target after a 5\,cm push and fail
on a UR5e, and a replay can only grasp as recorded. \method{} keeps only the effect, as a program of object keypoints, a terminal state
and two contact points (bottom), and re-solves the execution from the live scene: the same episode succeeds, the
push is absorbed, a UR5e runs it without retraining, and rotating the contact points turns the grasp.}
\label{fig:teaser}
\vspace{-6pt}
\end{figure}

As illustrated in Figure~\ref{fig:teaser}, our key observation is that the change in the world is the task,
and the demonstrator's motion is one of many ways to produce it. We therefore compile a demonstration into an
\emph{effect program}: keypoints of each relevant object tracked over time, two contact points recording where
the object was held, and the configuration the objects end in. Nothing in the program refers to the robot.
Based on this view we propose \method{}, a world-action model that describes effect, end-effector execution,
native action and terminal state as four streams with separate flow-matching times
\citep{unidiffuser,flowmatching,uwm}, so that every conditional is a query of the same weights. Clamping the program and sampling the execution poses the question of how to make the effect happen, which turns inference into programming. At deployment the unfinished part of the program is re-solved from the live scene every few
control steps. Because a push moves an object just as a grasp does, progress through the program is counted
only for effects the robot has caused, and a disturbance becomes a new observation.

We evaluate \method{} on LIBERO, Meta-World and RLBench, on four arms, under pushes, from pixels and on a real
Franka arm programmed by human videos. Where the demonstrator's motion stays valid, copying it is stronger; the
program pays off where copying breaks: after a push, on another arm, and from a person's hand.

We summarize our contributions as follows: (i)~we formulate the task carried by a demonstration as an
actor-free effect program of object keypoints, contact points and a terminal state, and propose \method{}, a world-action model whose independently noised streams let one set of weights both execute and predict; (ii)~we show that the
program is an interface a person can edit and hand to another body: its terminal state moves the placement, its contact points turn the grasp, and one program runs on four simulated arms and, from a human video, on a real arm; (iii)~we treat progress as an effect the robot causes and gate the program on engagement, so that
re-solving absorbs pushes that defeat replay; and (iv)~we show that editability needs contact written as points,
not weights, and training data with several executions of one effect.

\section{Related Work}
\label{sec:related}
\subsection{World-action models and imitation from one demonstration}
World-action models learn video and action jointly, so that one network can both predict and act. UWM
\citep{uwm} couples video and action diffusion with independent timesteps, JOPAT \citep{jopat} adds a point-track
stream, and Hydra-0 and Zero-WAM \mbox{\citep{hydra0,zerowam}} condition multi-billion-parameter video models on
flow or on a human video; generalist policies and pretrained world-action models specify the task with a sentence
\citep{openvla,openwam}. Methods that imitate one demonstration at test time read it as point-cloud graphs
\citep{instantpolicy}, as keypoint and action tokens \citep{kat}, through demonstration-to-scene correspondences
\citep{matchingpolicy} or as the video itself \citep{osviwm,vivla}, or they retarget the demonstrator's trajectory
\citep{demodiffusion}; each carries the demonstrator's motion. We keep independent timesteps but condition on an
actor-free program that a person can read and edit.

\subsection{Object motion as the task interface}
Several methods drop the actor and pass the object's motion to the robot. Most generate that motion, as 2D or 3D
flow from language, images, point clouds or generated video
\citep{im2flow2act,track2act,generalflow,avdc,partflow,novaflow,dream2flow} or as a trajectory from a per-task
model \citep{spot,motionfield3d}, and execute it with a flow-conditioned policy, a controller or trajectory optimisation. Others reproduce the motion in one human video by optimisation, retargeting or a policy trained for
that demonstration \citep{orion,rsrd,human2sim2robot}. Keypoints and trajectories also serve as task specifications
\citep{kpam,rekep,transporter,rttrajectory}, ATM and Point Policy \citep{atm,pointpolicy} condition on point tracks
\citep{cotracker} that include the actor's motion, and demonstration editing adapts recorded demonstrations to new
layouts to synthesise data \citep{mimicgen,demogen}. We share the premise that the object's motion is the task.
Closest to us, Im2Flow2Act \citep{im2flow2act} also carries a task from human video to a robot through object
flow, but generates the flow with a separate model and executes it with a flow-conditioned policy; we read the
effect from the demonstration itself, add contact points and a terminal state, and sample the execution from one
multi-task world-action model that also answers forward queries of the same program.

\section{Method}
\label{sec:method}

\begin{figure}[!tbp]
\centering
\includegraphics[width=\textwidth]{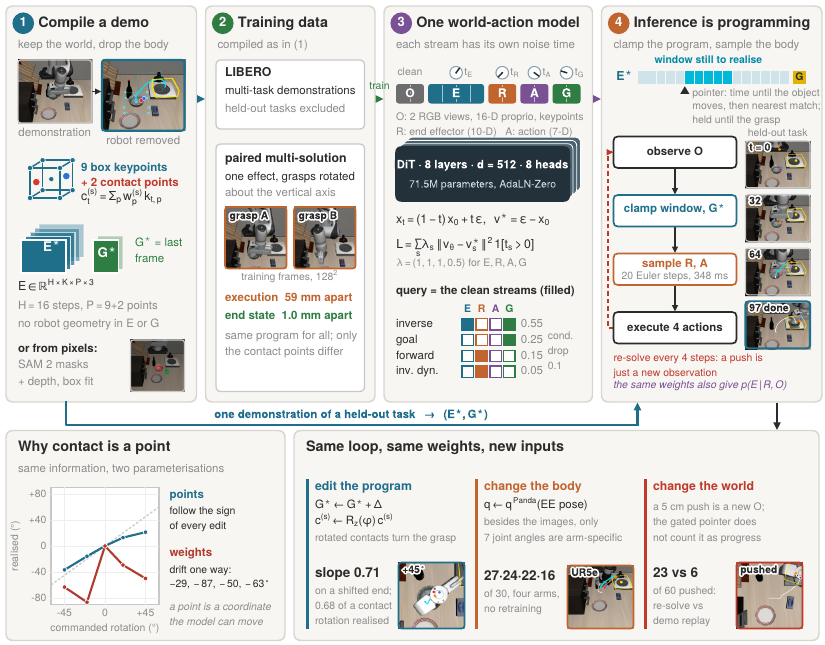}
\caption{\textbf{\method{} end to end.} (1) Compiling a demonstration into an effect program (\S\ref{sec:program}); (2) the training data; (3) one model with a flow-matching time per stream (\S\ref{sec:model}); (4) the closed loop of Algorithm~\ref{alg:loop}. Bottom: contact as points, and three operators of the same loop.}
\label{fig:pipeline}
\end{figure}

\subsection{From a demonstration to an effect program}
\label{sec:program}
The observation $O_t$ holds two RGB views, proprioception and the entities' current keypoints as an
anchor. For $K$ entities, the \emph{effect} $E \in \mathbb{R}^{H \times K \times P \times 3}$ is the
world-frame trajectory, over a horizon $H{=}16$, of $P$ points per entity: the eight corners and the centre
of its oriented box, plus two \emph{contact points} $c^{(1)}, c^{(2)}$ at the centroids of the two
finger-side contact patches at the moment of grasp. Each contact point is a fixed affine combination of the
entity's nine keypoints, $c^{(s)}_t = \sum_{p=1}^{9} w^{(s)}_p\, k_{t,p}$ with $\sum_p w^{(s)}_p{=}1$, so it
moves rigidly with the object; an entity the gripper never touches gets its own centre. The contact points
bring the grasp into the program: rotating them asks for another grasp of the same effect. The \emph{terminal
state} $G \in \mathbb{R}^{K \times P \times 3}$ is the last frame of the episode's effect. The
\emph{execution} $R \in \mathbb{R}^{H \times 10}$ holds end-effector position, 6-D rotation and gripper, and
the \emph{action} $A \in \mathbb{R}^{H \times 7}$ is the robot's native operational-space command. $E$ and $G$ contain no robot geometry; $R$ and $A$ describe only the body.

A demonstration of the target task yields $(\Estar,\Gstar)$ by reading entity poses along the recording. To
use a program in a new layout, each entity's trajectory is translated to that entity's present position. In
simulation the entity poses are read from the simulator, for the program and for the live keypoints in $O_t$
that the pointer of \S\ref{sec:infer} matches against it; from pixels they come from SAM~2 masks \citep{sam2}
on RGB-D, rendered or recorded by a real camera (\S\ref{sec:real}).

\subsection{One model, four streams}
\label{sec:model}
Tokens $[\,O \mid E\ (H{\times}K) \mid R\ (H) \mid A\ (H) \mid G\ (K) \mid \text{registers}\,]$ are
processed by one DiT \citep{dit} (8 layers, $d{=}512$, 8 heads, 71.5M parameters) with AdaLN-Zero
modulation. Observation tokens are never noised. Each stream $s$ draws its own flow-matching time
$t_s\in[0,1]$ per example, with $x_{t} = (1-t)\,x_0 + t\,\epsilon$ and target $v^\star = \epsilon - x_0$:
\begin{equation}
\mathcal{L} \;=\; \sum_{s \in \{E,R,A,G\}} \lambda_s \,
\mathbb{E}\Big[\big\| v_\theta(x_{t}, t)_s - v^\star_s \big\|^2 \cdot \mathbf{1}[t_s > 0]\Big],
\qquad \lambda = (1, 1, 1, 0.5).
\label{eq:loss}
\end{equation}
A stream with $t_s{=}0$ is given clean and contributes no loss, so the set of clean streams defines a
\emph{query}. One set of weights thus answers every conditional; each training example draws one: \emph{inverse} ($E, G$ clean; $R, A$ sampled), \emph{goal} ($G$ clean),
\emph{forward} ($R$ clean; $E, G$ sampled) or \emph{inverse dynamics} ($R$ clean; $A$ sampled).

\paragraph{Training data.} The model trains on LIBERO \citep{libero} tasks outside the held-out set. A paired
\emph{multi-solution} corpus adds executions that realise one effect through grasps rotated about the
object's vertical axis; solutions of one effect differ by a median 59\,mm in execution and 1.0\,mm in
terminal state. Every solution is conditioned on the program its group shares, so the contact points are the only input
that tells the solutions of one effect apart.

\subsection{Inference is programming}
\label{sec:infer}
\paragraph{A query is a clamp.} A query fixes a set $\mathcal{C}$ of streams to clean values with $t_s{=}0$.
Every other stream starts from $x_s\sim\mathcal{N}(0,I)$ at $t{=}1$ and is integrated to $t{=}0$ with $N{=}20$
Euler steps,
\begin{equation}
x_s \leftarrow x_s - \tfrac{1}{N}\, v_\theta(x, t)_s, \qquad s\notin\mathcal{C}.
\label{eq:sample}
\end{equation}
Executing a program is the inverse query, $\mathcal{C}{=}\{E,G\}$, with $E$ set to the window
$\Estar_{j+1:j+H}$ that a pointer $j$ selects and $G{=}\Gstar$; one call takes 348\,ms on one RTX~A6000. The
first $K_{\text{exec}}{=}4$ sampled actions run, the scene is observed again, and the loop repeats
(Algorithm~\ref{alg:loop}).

\paragraph{The pointer.} Let $k_t$ be the live keypoints of the moved entities, $\Estar_i$ the programmed
ones at step $i$ of a program of length $T$, and $j_{\text{move}}$ the first step at which a programmed entity
has moved 1\,cm. Matching is monotone within 40 steps,
\begin{equation}
\hat{j}_t = \operatorname*{arg\,min}_{j_{t-1}\le i< j_{t-1}+40} \big\lVert \Estar_i - k_t \big\rVert,
\label{eq:pointer}
\end{equation}
and $j_t{=}\hat j_t$ once $\hat j_t\ge j_{\text{move}}$. Before the first programmed motion the pointer follows
elapsed time but stays $H/2$ short of it, $j_t{=}\min\big(\max(\hat j_t,t),\max(\hat j_t, j_{\text{move}}{-}H/2)\big)$.

\paragraph{Progress is an effect the robot causes.} A push moves an object just as a grasp does. Engagement $e_t$ switches to 1 at
the first step at which the gripper is commanded closed and the moved entities have travelled 1\,cm since it
closed. While $e_t{=}0$, a match at or past $j_{\text{move}}$ is discarded ($\hat j_t\leftarrow j_{t-1}$), and
each entity's window is re-anchored to where the entity is, by an offset that decays to zero at $\Gstar$:
\begin{equation}
\tilde E_i = \Estar_i + \Big(1-\tfrac{i-j_t}{T-1-j_t}\Big)\big(k_t - \Estar_{j_t}\big), \qquad j_t< i\le j_t{+}H.
\label{eq:reanchor}
\end{equation}
At engagement the pointer jumps to the program's own engagement frame, $j_t\leftarrow\max(j_t, j_{\text{engage}})$,
found in the demonstration by the same rule, so the program's grasp and the scene's grasp coincide.

\paragraph{Editing and changing the body.} An edit changes the program's coordinates: $\Gstar\leftarrow\Gstar+\Delta$,
with the trajectory blended linearly toward the shift from the first motion, or
$c^{(s)}\leftarrow R_z(\varphi)\,c^{(s)}$ about the entity's vertical axis, which turns the grasp rigidly.
Apart from the camera images, only the seven joint angles $q$ in $O_t$ are specific to the arm. On another arm
they are re-expressed as the training arm's joints at the same end-effector pose,
$q\leftarrow \mathrm{IK}_{\text{Panda}}\big(\mathrm{FK}(q)\big)$; on the training arm this is the identity, and no
weights change.

\begin{algorithm}[t]
\caption{Closed-loop programming}
\label{alg:loop}
\small
\begin{algorithmic}[1]
\Require program $(\Estar,\Gstar)$, weights $\theta$, $H{=}16$, $K_{\text{exec}}{=}4$;\; $j\gets 0$, $e\gets 0$
\While{the episode runs}  \State observe $O_t$ and $k_t$; update $j$ (Eq.~\ref{eq:pointer}), discarding matches past $j_{\text{move}}$ while $e{=}0$
  \State $E\gets \Estar_{j+1:j+H}$;\; \textbf{if} $e{=}0$ \textbf{then} re-anchor $E$ (Eq.~\ref{eq:reanchor})
  \State sample $(R,A)$ with $E$ and $G{=}\Gstar$ clamped (Eq.~\ref{eq:sample})
  \State execute $A_{1:K_{\text{exec}}}$; at engagement set $e\gets 1$ and $j\gets\max(j, j_{\text{engage}})$
\EndWhile
\end{algorithmic}
\end{algorithm}

\section{Experiments}
\label{sec:exp}

\paragraph{Benchmarks and protocol.} We evaluate on held-out tasks and objects of LIBERO, on the test classes of
Meta-World ML10 and ML45, on RLBench, on four robot arms and on a real Franka arm programmed by human video
(Table~\ref{tab:setup}); simulated rollouts use the benchmark's own success check. LIBERO episodes re-plan every $K_{\text{exec}}{=}4$ steps under a 400-step cap, with one
fixed environment and sampling seed and 30 episodes per task and cell. As for the published baselines, a
LIBERO episode starts where the demonstration that programs it started; on Meta-World each episode is programmed
by another episode's demonstration. Outside \S\ref{sec:real}, keypoints come from simulator state. Counts are
from one training; replicates move the LIBERO held-out sum by up to seven (Appendix~\ref{app:diag}).

\subsection{One demonstration programs a held-out task}
\label{sec:main}\label{sec:beyond}

\begin{table}[t]
\centering\footnotesize
\caption{\textbf{Held-out tasks from one demonstration}, success rate (\%). Left: the three held-out LIBERO-Goal
tasks, 90 episodes; each published method reads the same demonstration through its own interface: UWM its video
frames, Instant Policy its point clouds, ATM its point tracks and Zero-WAM its video; the reference rows replay
its actions or give the same weights its end-effector trajectory (Appendix~\ref{app:uwm}). Right: the
Meta-World test classes selected by the rule of Appendix~\ref{app:prereg}, from ML10 (shelf-place,
sweep-into) and ML45 (bin-picking; shelf, sweep and bin in the table); \method{} is the mean of three trainings, each over 30 episodes programmed by another episode's
demonstration; published policies are fine-tuned on ten demonstrations of the class \citep{dami}. On the
five-class mean \method{} is below DAMI (49.8 against 56.8 on ML10, 34.0 against 37.5 on ML45;
Appendix~\ref{app:metaworld}).}
\label{tab:main}
\begin{minipage}[t]{0.45\textwidth}\vspace{0pt}\centering
\setlength{\tabcolsep}{2.6pt}\renewcommand{\arraystretch}{1.2}
\begin{tabular}{@{}lc@{}}
\toprule
\multicolumn{2}{@{}l}{\emph{LIBERO, held-out tasks}} \\
method & success \\
\midrule
UWM \citep{uwm} & 21.1 \\
Instant Policy \citep{instantpolicy} & \phantom{0}5.6 \\
ATM \citep{atm} & \phantom{0}0.0 \\
Zero-WAM \citep{zerowam} & \phantom{0}0.0 \\
\midrule
\method{} (ours) & \best{44.4} \\
\midrule
\multicolumn{2}{@{}l}{\emph{reference, carries the demonstrator's motion}} \\
replay of the demonstration & 96.7 \\
\method{}, demonstrator's execution & 94.4 \\
\bottomrule
\end{tabular}
\end{minipage}\hfill
\begin{minipage}[t]{0.53\textwidth}\vspace{0pt}\centering
\setlength{\tabcolsep}{2.6pt}\renewcommand{\arraystretch}{1.2}
\begin{tabular}{@{}lccc@{}}
\toprule
\multicolumn{4}{@{}l}{\emph{Meta-World, held-out classes}} \\
method & shelf & sweep & bin \\
\midrule
Meta-World baseline \citep{metaworld} & \phantom{0}0.0 & \phantom{0}24.7 & \phantom{0}0.3 \\
DP3 \citep{dp3} & 12.7 & \phantom{0}16.7 & \phantom{0}9.3 \\
Mamba Policy \citep{mambapolicy} & 10.0 & \phantom{00}9.7 & \phantom{0}9.7 \\
FreqPolicy \citep{freqpolicy} & \phantom{0}7.0 & \phantom{0}15.7 & 11.0 \\
FlowPolicy \citep{flowpolicy} & \phantom{0}0.0 & \phantom{00}6.7 & \phantom{0}9.0 \\
DAMI \citep{dami} & 10.7 & \phantom{0}28.0 & 13.7 \\
\midrule
\method{} (ours) & \best{22.2} & \best{100.0} & \best{52.2} \\
\bottomrule
\end{tabular}
\end{minipage}

\end{table}

\begin{figure}[t]
\centering
\includegraphics[width=\textwidth]{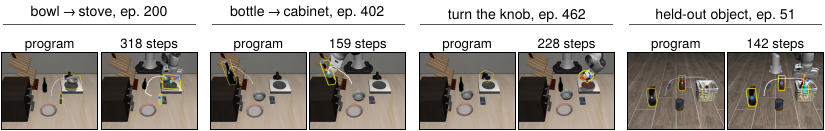}
\caption{\textbf{One program per held-out task, executed closed loop.} Left of each pair: the program on its
demonstration with the robot removed (object track white, terminal state $\Gstar$ yellow, contact points
red and blue). Right: the last frame of the execution, with the object's path in white and its length in steps above. The last pair is a LIBERO-Object
episode with a held-out object (the sauce bottle into the basket). Episodes are the lowest-numbered success of each
task in its held-out evaluation; failures and intermediate frames are in Appendix~\ref{app:gallery}, and the baselines on the same episodes in Appendix Figure~\ref{fig:baselines}.}
\label{fig:heldout}

\end{figure}

Table~\ref{tab:main} gives the central result. The three held-out LIBERO-Goal tasks (bowl
on the stove, bottle on the cabinet, stove knob) contribute no robot data to training. Given one demonstration of such a task, \method{} completes 40 of 90 episodes from the full program and 49
from its terminal state alone, while the same backbone and data given a goal image, language or no
specification complete none (Appendix~\ref{app:diag}); the task interface is the only difference. Most of the 40 come from the bowl, whose nearest training effect lies 4.2\,cm away. That the terminal state alone does better shows that success here comes mainly from an object-frame goal; the trajectory and contacts matter for what \S\ref{sec:edit} edits and \S\ref{sec:cross} carries. Since each episode starts where its demonstration did, replaying the demonstration completes 87 of 90; the program is meant to keep working where replay does not (\S\ref{sec:cross},~\ref{sec:real}). In the rollouts (Figure~\ref{fig:heldout}) successful bowls and bottles end a median 2.1 and 1.6\,cm from the programmed destination. On a second split with three other held-out tasks, the
program completes 27 of 90 (Appendix~\ref{app:fold2}). With the bowl moved 4 or 8\,cm from its demonstrated start, the program completes 12 of 15 and 10 of 15 episodes, against 27 of 30 from the demonstrated start (Appendix~\ref{app:diag}).

\paragraph{Published methods on the same episodes.} All successes of UWM and Instant Policy are on the knob (Instant Policy reaches 7 with two demonstrations),
and none of the published methods completes a pick-and-place episode: in the recordings they hover, stall at the
object or move elsewhere, and Zero-WAM moves the arm but never the target object (Appendix~\ref{app:baselines}). Each reads the demonstration as the arm's motion or as
pixels, the two forms the program replaces with the effect alone. The rows measure released methods, not their ceiling: Instant Policy and Zero-WAM run zero-shot and ATM within a fixed budget.

\paragraph{Objects the model never manipulated.} LIBERO-Object applies one skill to ten objects. We remove a bottle, a box
and a carton from training and train the \method{} recipe unchanged. Programs carry over to the removed
objects (Appendix Table~\ref{tab:object} and Figure~\ref{fig:object}): 67 of 90 from the program (65 in a second training) and 89 from the terminal state. Trained without LIBERO-Object, the same recipe completes none and leaves the object where it started.

\paragraph{Held-out classes chosen by the benchmark.} Meta-World \citep{metaworld} fixes its own splits:
ML10 and ML45 train on 10 or 45 classes and test on five classes chosen by the benchmark's authors, on a
Sawyer arm with scripted-expert demonstrations. We train the \method{} recipe on each training set and
evaluate every test class under a layout transfer, so no episode receives its own demonstration. On each
split we report the recipe with the highest five-class sum under the transfer: the effect written relative
to each entity's anchor on ML10, and a joint-position channel for articulated entities on ML45; \S\ref{sec:analysis}
explains why the two splits prefer different coordinates. On shelf-place, sweep-into and bin-picking the program exceeds every published result
(Table~\ref{tab:main}); the same backbone given no task specification completes none of these
classes. The comparison is per class and across protocols: on the five-class mean the program is below DAMI, every variant sees Meta-World's target marker, and the unconditioned reference beats every published row on two ML45 classes (Appendix~\ref{app:metaworld}).

\paragraph{Twenty-four held-out tasks on RLBench.} RLBench \citep{rlbench} changes the simulator, the robot
interface and the source of demonstrations at once, and we hold out all 24 tasks of the Instant Policy
evaluation. With the action written as an end-effector increment, the program completes 160 of 720 episodes
against 108 for the same recipe given no program, far below Instant Policy, which completes 575 with one
demonstration (Appendix~\ref{app:rlbench}).

\subsection{The program is editable}
\label{sec:edit}

\begin{figure}[!tbp]
\centering
\includegraphics[width=\textwidth]{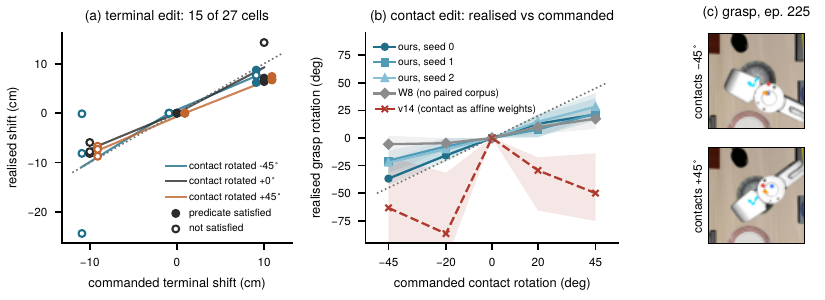}
\caption{\textbf{Editing the program} (bowl task). (a) Terminal edits on a $3\times3$ grid of contact
rotation and terminal shift: realised against commanded shift, filled where the task predicate fires. (b)
Contact edits: realised grasp rotation against commanded rotation of the two contact points, median of
episode-paired shifts with 95\% bootstrap bands. Contacts stored as affine weights (red) are read with the wrong sign. (c) The grasp with contacts rotated by $\mp45^\circ$, from above.}
\label{fig:edit}

\end{figure}

\paragraph{Where the object ends.} Translating $\Gstar$ moves the realised placement with it
(Figure~\ref{fig:edit}a): over 30 episodes shifted by $\pm5$ and $\pm10$\,cm the slope is 0.71.

\paragraph{How the object is held.} Rotating the two contact points about the object's vertical axis turns the grasp the arm takes, measured from simulator contacts
(Figure~\ref{fig:edit}b,c). The grasp turns in the commanded direction at every rotation and realises on average 0.68 of it, and with edited contacts the model completes 22 to 24 of 30 at $\pm20^\circ$ and $\pm45^\circ$, against 27 unedited
(Appendix~\ref{app:editing}).

\subsection{One effect, many bodies and executions}
\label{sec:cross}

\begin{figure}[!tbp]
\centering
\begin{subfigure}[b]{0.50\textwidth}
\includegraphics[width=\textwidth]{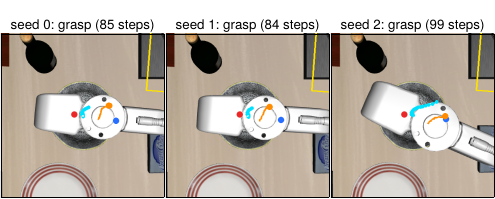}
\caption{Three sampling seeds, one program (episode 224).}
\end{subfigure}\hfill
\begin{subfigure}[b]{0.47\textwidth}
\includegraphics[width=\textwidth]{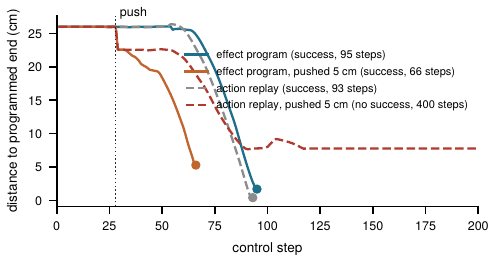}
\caption{A 5\,cm push at step 28 (episode 206).}
\end{subfigure}
\caption{\textbf{Execution is a free variable.} (a) One effect, three grasps: seed 2 takes the bowl from the
other side, and all three succeed. (b) The bowl's distance to its programmed end. Re-solving the remaining
program absorbs the push; replaying the demonstration's actions ends 8.0\,cm short.}
\label{fig:free}
\end{figure}

\begin{table}[t]
\centering\footnotesize
\caption{\textbf{One program, four arms, no retraining}: successes of 30 on the bowl / bottle tasks for the
first training of each recipe, factored over the contact channel and the paired multi-solution corpus.
Proprioception is re-expressed in the training arm's joint space, the identity on the Panda. Mean: sums
averaged over trainings (two per recipe, three for \method{}); every training is in Table~\ref{tab:arms_full}.}
\label{tab:arms}
\setlength{\tabcolsep}{3.1pt}
\begin{tabular}{@{}llrrrrrrrr@{}}
\toprule
 & & \multicolumn{4}{c}{first training, bowl / bottle} & \multicolumn{2}{c}{sum} & \multicolumn{2}{c}{mean sum} \\
\cmidrule(lr){3-6}\cmidrule(lr){7-8}\cmidrule(l){9-10}
ingredients & contact / corpus & Panda & IIWA & Kinova3 & UR5e & bowl & bottle & bowl & bottle \\
\midrule
neither          & --- / ---   & 28/18 & 26/7  & 17/16 & 8/0  & 79 & 41 & 78.5 & 41.0 \\
contact only     & yes / ---   & 27/17 & 21/13 & 24/10 & 2/1  & 74 & 41 & 82.0 & \best{48.5} \\
corpus only      & --- / yes   & 24/7  & 15/0  & 11/2  & 14/1 & 64 & 10 & 63.5 & 12.0 \\
both (\method{}) & yes / yes   & 27/15 & 24/17 & 22/5  & 16/6 & \best{89} & \best{43} & \best{83.3} & 27.7 \\
\bottomrule
\end{tabular}
\end{table}

\paragraph{Four arms, one program.} Table~\ref{tab:arms} executes it on the Panda it was trained with and on a KUKA IIWA, a Kinova Gen3 and
a UR5e (Appendix Figure~\ref{fig:arms}). \method{} completes 27, 24, 22 and 16 of 30 bowl
episodes. The transfer rests on re-expressing proprioception in the Panda's joint space and on the
operational-space controller the four arms share: given each robot's own joint angles instead, the new robots
complete 1, 0 and 0. The table also crosses the contact channel with the paired corpus (\S\ref{sec:analysis});
on the bottle the corpus lowers the mean sum (27.7 with both ingredients, against 48.5 with the contact channel alone).

\paragraph{Several executions of one effect.} Three sampling seeds succeed in 27, 28 and 28 of 30 through grasps a median $20.1^\circ$ apart, and at least one seed succeeds in all 30
(Figure~\ref{fig:free}a).

\begin{wraptable}{r}{0.5\textwidth}
\vspace{-12pt}
\centering\footnotesize
\caption{\textbf{A 5\,cm push}: successes of 30, undisturbed / pushed at step 28, and the pushed sum over both
tasks with the push at step 32. All cells are in Table~\ref{tab:disturb}.}
\label{tab:disturb_main}
\setlength{\tabcolsep}{2.4pt}
\begin{tabular}{@{}lrrr@{}}
\toprule
 & bowl & bottle & $\Sigma$, step 32 \\
\midrule
replay of the demonstration & 29 / 6 & 28 / 0 & 9 \\
\method{} & 27 / \best{9} & 12 / \best{14} & \best{28} \\
\quad without the gate & 26 / 3 & 13 / 4 & 10 \\
\bottomrule
\end{tabular}
\vspace{-8pt}
\end{wraptable}
\paragraph{A world that does not cooperate.} We push the bowl or the bottle 5\,cm mid-task and compare
re-solving with open-loop replay of the demonstration's actions (Table~\ref{tab:disturb_main},
Figure~\ref{fig:free}b). Re-solving completes 23 of 60 pushed episodes and replay 6. Without the engagement
gate of \S\ref{sec:infer}, the pointer reads the push as progress, jumps past the grasp, and re-solving falls to 7 (Appendix~\ref{app:disturb}).

\subsection{From pixels to a real arm}
\label{sec:real}

\begin{figure}[!tbp]
\centering
\includegraphics[width=0.72\textwidth]{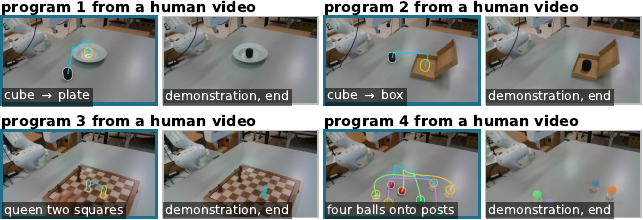}
\caption{\textbf{Programs compiled from real human videos.} Each program is drawn on the demonstration's
first frame, before the hand enters (white: entities at the start; one colour per entity for its route;
yellow: its terminal region $\Gstar$), beside the demonstration's last frame. Programs 1 and 2 share a scene:
the same cube carried to a plate and to a box.}
\label{fig:real_pair}
\end{figure}

\begin{table}[t]
\centering\footnotesize
\caption{\textbf{A real Franka arm programmed by human videos}, successes of 10. Each program is compiled from
one human video; no robot demonstration of these tasks is used. Left: \method{} against the same backbone and training data conditioned on the video's last frame as a goal image. Right: \method{} on the cube-to-plate program as demonstrated (the trials of the left block), with its terminal state shifted by 10\,cm, and with the cube pushed 5\,cm mid-task.}
\label{tab:real}
\begin{minipage}[t]{0.52\textwidth}\vspace{0pt}\centering
\setlength{\tabcolsep}{4pt}\renewcommand{\arraystretch}{1.1}
\begin{tabular}{@{}lcc@{}}
\toprule
task & goal image & \method{} \\
\midrule
cube to plate & 8 & \best{10} \\
cube into box & 5 & \best{10} \\
cube onto tray & 5 & \best{8} \\
box into bin & 4 & \best{8} \\
\midrule
total (of 40) & 22 & \best{36} \\
\bottomrule
\end{tabular}
\end{minipage}\hfill
\begin{minipage}[t]{0.45\textwidth}\vspace{0pt}\centering
\setlength{\tabcolsep}{4pt}\renewcommand{\arraystretch}{1.1}
\begin{tabular}{@{}lc@{}}
\toprule
cube to plate & \method{} \\
\midrule
as demonstrated & 10 \\
terminal shifted left & 9 \\
terminal shifted right & 9 \\
pushed 5\,cm mid-task & 7 \\
\bottomrule
\end{tabular}
\end{minipage}
\end{table}

\paragraph{Rendered RGB-D.} Fitting oriented boxes to SAM~2 masks of rendered RGB-D gives a program without
simulator state; in the episode of Appendix Figure~\ref{fig:pixels} the bowl's centre is tracked to a median
0.61\,cm. Programs read this way complete 26 of 30 bowl and 19 of 30 bottle episodes, against 27 and 12 from simulator state; with the live keypoints segmented as well, so that nothing in the loop reads simulator state, they complete 40 of 60, against 39 (Appendix~\ref{app:live}). One centimetre of
keypoint noise leaves success unchanged (Appendix~\ref{app:noise}).

\paragraph{A real arm programmed by a person.} A person performs each task once at a table in front of a
Franka Research~3 while a calibrated RGB-D camera records. One click per entity on the first frame prompts
SAM~2 \citep{sam2}, which propagates the masks through the video, including hand occlusions; the masked depth, expressed in the robot's base frame, gives each entity's keypoints, and $\Gstar$ is read from the last frame; Figure~\ref{fig:real_pair} shows programs compiled the same way from RGB video. The arm then executes the program closed
loop, with live keypoints from the same camera (Appendix~\ref{app:real}). The model is the simulation
checkpoint fine-tuned on 156 real demonstrations of other tasks, none of the four evaluated;
the goal-image variant shares that data. Over four tasks \method{} completes
36 of 40 trials, against 22 for the same backbone conditioned on the video's last frame as a goal image
(Table~\ref{tab:real}). With the terminal state shifted 10\,cm to either side, the arm completes 9 and 9 of 10, and after a 5\,cm push 7 of 10. Ten further programs are in Appendix~\ref{app:real}.

\subsection{What makes a program programmable}
\label{sec:analysis}

\begin{wraptable}{r}{0.5\textwidth}
\vspace{-12pt}
\centering\scriptsize
\caption{\textbf{Core ablations}: one component changed per row, on the measure it acts on. All variants: Table~\ref{tab:ablations}.}
\label{tab:ablate_core}
\setlength{\tabcolsep}{2.4pt}
\begin{tabular}{@{}llrr@{}}
\toprule
changed & measure & \method{} & variant \\
\midrule
no contact channel & four arms, bowl / bottle & 89 / 43 & 64 / 10 \\
no paired corpus & contact following & 0.68 & 0.31 \\
demo-clock pointer & held-out bottle / $\Sigma$ & 12 / 40 & 1 / 27 \\
native proprioception & four arms, bowl & 89 & 28 \\
relative effect & held-out $\Sigma$ & 40 & 1 \\
\bottomrule
\end{tabular}
\vspace{2pt}
\end{wraptable}
\paragraph{Points, not weights.} A model trained on the nine affine weights that define the contact points, the same information, completes 26 of 30 unedited bowl episodes and a cross-arm bowl sum of 85. Yet whatever the sign of the commanded
rotation, its realised offset is negative ($-29$, $-87$, $-50$, $-63^\circ$; Figure~\ref{fig:edit}b, red),
and the sign error repeats in a retraining on a second machine and with a second training seed
(Appendix~\ref{app:v100}).

\paragraph{Diversity of execution buys contact editability.} Without the paired corpus, the model follows contact edits less closely: at 50 episodes per rotation, all four trainings with the corpus follow more closely than all three without (0.41 to 0.70 against 0.31 to 0.34; one-sided exact permutation $p{=}0.029$, Appendix~\ref{app:diag}). On the four robots the two ingredients interact: with both, the mean bowl sum is 16 above what their separate gains predict (Table~\ref{tab:arms}).

\paragraph{Absolute or relative coordinates.} Written as displacements, the program completes 1 of 90 held-out LIBERO episodes but opens the ML10 drawer in 16 of 30 where absolute coordinates open none: a LIBERO effect ends at a \emph{place}, the drawer's is a \emph{motion} along its slide. The same weights also predict the effect of an execution (Appendix Figure~\ref{fig:forward}).

\section{Limitations and conclusion}
\label{sec:limit}
Effects that barely move an object, such as turning a knob, leave the program little to specify; the
joint-position channel of Meta-World is the natural extension (Appendix~\ref{app:limits}). The pointer and
gate are hand-set rules for grasped objects; entities are rigid boxes, mostly read from simulator state; where
the demonstrator's motion stays valid, copying it is stronger; RLBench's specialist remains far ahead; and most
cells are single trainings of 30 episodes, whose replicates can move by as much as the differences we discuss.

Keeping only what a demonstration did to the world gives a world-action model an input that people can edit and that carries across bodies, and reading progress from caused effects lets it absorb a push.

\bigskip
\subsubsection*{Ethics statement}
The closed-loop experiments use the public LIBERO, RoboCasa, Meta-World and RLBench benchmarks and their released or scripted demonstrations, and trials on a Franka arm in our lab. The real videos of Appendix~\ref{app:real} were recorded in our lab and
show the hands and forearms of lab members performing table-top tasks; no faces or other personal data
appear in the frames shown. The baselines were run from their authors' released code and checkpoints, and every
change made to run them on our protocol is stated with them. We see no direct path from this work to harm
beyond that of robot manipulation research.

\subsubsection*{Reproducibility statement}
Appendix~\ref{app:hp} gives the architecture, hyper-parameters, training data and schedule; the benchmark
appendices give each evaluation protocol, Appendix~\ref{app:baselines} the configuration of every baseline and
Appendix~\ref{app:real} the real-robot setup. Every simulation table cell is generated from per-episode evaluation records by one script rather than transcribed, and every simulation number in the text was re-derived from those records before submission. The real-robot counts of Table~\ref{tab:real} are tallied per trial against the criteria of Appendix~\ref{app:real}. The protocol is fixed in one place (re-plan every four steps, 400-step cap, one
environment seed, 30 episodes per cell), and any departure is named in the caption. One training takes about
3.5 hours on three consumer GPUs. Upon acceptance we will release all code, model weights, the corpus
construction, the evaluation scripts and the pre-registration files (Appendix~\ref{app:prereg}).

\subsubsection*{Use of large language models}
We used large language model assistants for editing (grammar, spelling, and word choice). All AI-assisted text was reviewed by the authors, and we take responsibility for the final content of this work.

\FloatBarrier
\bibliography{iclwam}
\bibliographystyle{iclr2027_conference}

\appendix

\section{Implementation details}
\label{app:hp}
\paragraph{Inputs and queries.} Each observation contains two $128^2$ RGB views (the agent view and the wrist
camera), 16-D proprioception (end-effector position, orientation and gripper state, plus the arm's seven joint
angles), and the entities' current keypoints, which serve as an anchor. During training each example is posed as
one of four queries, drawn with probabilities 0.55 (inverse), 0.25 (goal), 0.15 (forward) and 0.05 (inverse
dynamics), and conditions are dropped with probability 0.1.

\paragraph{Model and training.} All streams are processed by one shared DiT backbone with depth 8, width 512,
8 heads and 8 register tokens (71.5M parameters). A per-view ResNet-18 with GroupNorm encodes the observation,
and the entity keypoints at $t{=}0$ enter as an anchor token. The effect is written in absolute normalised
world coordinates, and the stream losses are weighted $E{:}R{:}A{:}G = 1{:}1{:}1{:}0.5$. The inverse query gets
extra regularisation: $G$ is given clean with probability 0.8, the condition dropout of 0.1 applies here too,
and images are randomly cropped by 4 pixels. We use batch 64 per rank on three ranks and AdamW (learning rate
$3\times10^{-4}$, weight decay $10^{-4}$, $\epsilon{=}10^{-6}$), with 1000 warm-up steps, gradient clipping at
1.0 and EMA 0.999. Training runs in fp32 because bf16 autocast diverged twice between 32k and 33k steps. The
learning rate follows a 60k cosine schedule that we stop at 30k, a point the longest run, from 20k to 60k, supports with bowl-task sums over the four robot arms of 81, 74, 73, 76 and 66.
One training run takes about three and a half hours on three RTX~3090s (195 and 221 minutes for the two
replicates we timed).

\paragraph{Training data.} The LIBERO data covers 107 tasks (LIBERO-90, LIBERO-10 and the LIBERO-Goal tasks
outside the held-out set), which yield 1.16M training windows. The paired multi-solution corpus adds 56 scenes
with 273 executions.

\begin{table}[h]
\centering\small
\caption{\textbf{Experimental setup.} Each test episode is a closed-loop rollout, scored by the benchmark's own
success check in simulation and by the criterion of Appendix~\ref{app:real} on the real arm. Every held-out episode is programmed by a single demonstration, and no parameters are updated at test time.}
\label{tab:setup}
\setlength{\tabcolsep}{3pt}\scriptsize
\begin{tabular}{@{}p{2.55cm}p{2.3cm}p{2.45cm}p{1.35cm}p{4.3cm}@{}}
\toprule
split & simulator, robot & held out at test & test episodes & compared with \\
\midrule
LIBERO-Goal (\S\ref{sec:main}) & MuJoCo, Panda & 3 tasks (new goals) & $3\times30$ & other task interfaces; UWM, Instant Policy, ATM, Zero-WAM \\
LIBERO-Object (\S\ref{sec:beyond}) & MuJoCo, Panda & 3 objects & $3\times30$ & same recipe without LIBERO-Object \\
Meta-World ML10 / ML45 (\S\ref{sec:beyond}) & MuJoCo, Sawyer & 5 classes each (benchmark-defined) & 30 per class & observation-only reference; published \\
RLBench (\S\ref{sec:beyond}) & CoppeliaSim, Panda & 24 tasks & $24\times30$ & observation-only reference; Instant Policy \\
four arms (\S\ref{sec:cross}) & MuJoCo, IIWA / Kinova3 / UR5e & the arm & 30 per task & ablations of contact and corpus \\
real robot (\S\ref{sec:real}) & lab table, Franka Research~3, RGB-D & 4 tasks from human videos & $4\times10$ & same backbone with a goal image \\
\bottomrule
\end{tabular}
\end{table}

\paragraph{Sampling.} We sample with 20 Euler steps and keep clamped tokens clean throughout. The control loop
re-plans every four control steps by calling the inverse query, so the cost of one re-plan sets the loop's
speed. On one RTX~A6000, excluding simulation, a re-plan takes a median of 348\,ms with 20 steps and
117\,ms with 5 (measured over 125 and 96 re-plans in two bowl episodes).

\paragraph{The paired multi-solution corpus.} The corpus gives the model several executions of the same
effect. To build it, we rotate the demonstrated grasp about the object's own vertical axis, which moves the
wrist but not the object, and replay each rotated grasp in the simulator. The replays do not reproduce the
object motion exactly, so conditioning each execution on its own program would let small program differences
stand in for the choice of grasp. We therefore condition every execution in a group on the group's canonical
program. This shrinks the difference between programs of one group from 28.3\,mm to 0.47\,mm (after phase
alignment) and from 12.5\,mm to zero at the terminal state, while the contact points still differ by
39.9\,mm. Within a group, the contact points are thus the only part of the program that tells the executions
apart. Across the two shards the headline model trains on (56 scenes, 273 executions), executions of one
effect differ by a median of 59\,mm and end in terminal states that differ by 1.0\,mm.

\section{Pre-registered criteria and outcomes}
\label{app:prereg}
For every architectural variant, the reading of its results was fixed in writing before any of its numbers
were seen. Each criteria file is under version control with the commit that introduced it and the date of that
commit, and that history will be released with the code upon acceptance. For example,
the criteria for v13 were committed on 09-07 at 07:58, and the run they judge was configured on 09-08 at 01:38;
for v14 the criteria were committed on 09-09 at 22:13 and the run was configured on 09-10 at 02:22. The
criteria reported in this paper and their outcomes are:

\begin{itemize}
\item \textbf{Contact as affine weights (v14).} Criteria: mean following over $\pm20^\circ,\pm45^\circ$ of at least 0.85, a held-out effect-program sum of at least 40 of 90, a $-45^\circ$ success of at least 24 of 30 and a cross-arm bowl sum of at least 81. Measured: a held-out sum of 44, a cross-arm bowl sum of 85, a $-45^\circ$ success of 13, and a realised offset of the same sign at every commanded rotation, the drift of \S\ref{sec:analysis}. The paper keeps contact as two points.
\item \textbf{Dense surface support (v13).} Pass marks: following 0.85, held-out sum 40, $-45^\circ$ success 24. Measured: 0.31, 29 and 14. The paper keeps two contact points.
\item \textbf{Axis-symmetric corpus (v15), gated before training.} Hypothesis: the direction asymmetry of contact following reflects how often each orientation occurs in the corpus. Gate: the density at the less-followed orientation is at most half the density at the other. The measured ratio is 1.10, so the corpus stays as it is. Measurements also cleared a modulo-180 read-out artefact, a yaw asymmetry in the corpus, the support band of training orientations and wrist joint limits; two further trainings show that the training seed sets the direction of the asymmetry (Appendix~\ref{app:editing}).
\item \textbf{Training past 30k steps (W8, 20k to 60k).} At least one checkpoint had to reach a four-arm bowl sum of 76; the three checkpoints the rule judges give 73, 76 and 66. The 50k checkpoint reaches 76, and the headline model stays at 30k, where every experiment of the paper runs.
\item \textbf{Ranking sampled executions (Go/No-Go 18).} One draw completes 26 and 19 episodes of tasks 4 and 8, and eight ranked draws 24 and 17, so the model executes its first draw (Appendix~\ref{app:editing}).
\item \textbf{The engagement gate.} Without disturbance, the held-out sum with the gate had to stay within 5
of 40, with at most five episodes changing outcome. Under pushes, re-solving had to complete at least 10 more
of 60 episodes than replay. Outcome: 40 with four changed episodes, and 23 against 6. A stricter gate (G2),
written after these numbers were seen, completed 29 pushed episodes but changed seven undisturbed outcomes,
so \S\ref{sec:infer} uses the first gate (Appendix~\ref{app:disturb}).
\item \textbf{A second held-out split.} The rule for drawing the split and its seed were fixed before the
draw; the pass mark was an effect-program sum of 20 of 90. Outcome: 27.
\item \textbf{A second training on LIBERO-Object.} A result at least 10 above the reference's 0 would count as
a reproduction. Outcome: 65.
\item \textbf{Retrained ablations R1--R4.} The bands were fixed before the first training: a held-out sum of
25 or less marks a component as needed for held-out execution, and a $\pm45^\circ$ following of 0.30 or less
marks it as carrying editability. All four variants fall inside both bands.
\item \textbf{Meta-World ML45 and ML10 (prereg 22 and 24).} Before any Meta-World number was read, the rule asked for a five-class sum 10 of 150 above the reference; the outcome is 48 against 47, so results are reported per class. The per-class rule of \S\ref{sec:beyond} was fixed before ML10 was read, and shelf-place, sweep-into and ML45's bin-picking pass it.
\item \textbf{The joint-state channel (prereg 25).} On LIBERO the rule asked for a gain of 10 of 30 on the knob task; the channel scores 0 against 1, so LIBERO keeps the headline variant. On Meta-World a row is reported whole: on ML45 the joint-state row has the higher five-class sum (34.0 against 26.7\% under the transfer) and is reported, and on ML10 the other row has (73 against 51 of 150).
\item \textbf{Effect relative to the anchor (prereg 26).} On LIBERO the rule asked for a gain of 10 of 30 on the knob task; the relative variant completes 0, 1 and 0 of 30 on the three held-out tasks, so LIBERO keeps absolute coordinates. On ML10, drawer-open and lever-pull together had to gain 10 of 60 under the transfer; they gain 11. The rule that picks each reported Meta-World row (the highest five-class sum under the transfer, reported whole) was written on 09-21 at 13:10, before any evaluation of the two new ML10 variants existed, and seeds 1 and 2 of both picked variants were trained after that choice. Amendment C, written before any RoboCasa evaluation of the two relative variants existed, was run on the earlier W3 recipe and set a line of 10 above that recipe's reference (4 of 90); the variants score 6 and 3.
\item \textbf{The reference given the program (unplanned).} This rule was written after the two sweep-into
evaluations had been seen and before the shelf-place ones. A result within 3 of the reported variant means the
program alone carries the class; a result 6 or more below means that training with the clamped program
contributes at least that much. Outcome: sweep-into 27 and 28 against 30 and 30; shelf-place 2 and 2 against
16 and 11 (Appendix~\ref{app:metaworld}).
\item \textbf{RLBench (prereg 23).} The program variant had to beat the reference by 48 of 720. With the absolute-pose head the margin was 16 for the absolute program and 41 for
the relative one. Amendment G, written before any conversion or training, retrains the three variants to
predict the end-effector increment, gated on replaying the held-out increments (G2: 110 of 117, against 108 of
117 for absolute-pose replay, after amendment G-2 moved the integration to the commanded setpoint). The margin
then became 52 for the effect program (160 against 108) and 15 for the relative one (Appendix~\ref{app:rlbench}). Amendment H, written before any full evaluation ran, adds a fallback for
poses the IK rejects and applies it to all three increment variants. It also counts the tasks on which the
program reaches Instant Policy (the larger of its published rate and our one-demonstration run). With the fallback the margin is 42 for the effect program (163 against 121) and 13 for the relative one; \S\ref{sec:beyond} reports the amendment-G configuration (Table~\ref{tab:rlbench_fb}).
\item \textbf{Controls before submission (prereg 27).} Written before any of its evaluations ran. Push at step
32: re-solving completes 28 of 60 against 9 for replay, the same ordering as for the push at step 28. Reversed
program starting from the end state: the line was 10 of 30 with a median distance below 5\,cm; the measured values are 2 of 30 and 25.7\,cm, so programs are edited through their terminal state and contact points (Table~\ref{tab:reverse_end}).
Terminal edits for the recipes without the contact channel or without the corpus: reported as measured
(Table~\ref{tab:terminal_arms}). Second seed of the weight-encoded model: the realised rotation had to be negative at all four commanded rotations. It is
($-86$, $-97$, $-84$ and $-96^\circ$ for $+20$, $-20$, $+45$ and $-45^\circ$). Second seeds of the $2\times2$
variants: the interaction $I = \text{both} - \text{contact only} - \text{corpus only} + \text{neither}$ on
the four-arm bowl sum, with \method{} averaged over its three trainings and the other variants over two, had to reach 10 to count as an interaction. It is 16.3 (83.3, 82.0, 63.5, 78.5), so \S\ref{sec:analysis} reports an interaction: the combined recipe has the highest bowl sum
(Tables~\ref{tab:arms} and \ref{tab:ablations}).
\item \textbf{Live keypoints from rendered RGB-D (prereg 28).} Written before any evaluation ran, with a smoke
test (median live centre error below 2\,cm, perception below 0.3\,s per step) gating the full evaluations.
The criterion: the pixel-only evaluation completes at least the bowl-plus-bottle sum of the simulator-state evaluation minus 10. It completed 40 against 39 (Appendix~\ref{app:live}), and \S\ref{sec:real} reports it.
\end{itemize}
\section{Scope}
\label{app:limits}
On LIBERO, the held-out tasks are new goals in known scenes and new objects for a known skill. Effects that
are small rotations of a fixed object, such as a stove knob, a lever or a drawer handle, give the program
little motion to specify; the joint-position channel for articulated entities that \method{} uses on
Meta-World ML45 extends the program to such effects. RoboCasa changes the kitchen and the mobile base every episode (Appendix~\ref{app:robocasa}). On RLBench the program adds 52 successes to the same recipe without a
program (Appendix~\ref{app:rlbench}).

\section{LIBERO: further analysis}
\label{app:diag}
\paragraph{Effects with little motion.} The LIBERO knob's centre moves 0.3\,cm over an episode. In the three
LIBERO-90 scenes that contain a knob the checkpoint completes it from its program 10 of 10 times, and in the
held-out scene 1 of 30. Drawer-open, lever-pull and door-unlock on Meta-World (Appendix~\ref{app:metaworld})
and the two drawer tasks of the second split (Appendix~\ref{app:fold2}) are effects of the same kind. With the
object displaced at the start, the program completes 12 of 15 episodes at 4\,cm and 10 of 15 at 8\,cm.

\begin{table}[h]
\centering\small
\caption{Held-out LIBERO-Goal tasks: successes out of 30 per task in every row. The per-task columns come
from the first training, which is the checkpoint every other experiment uses; the last column adds two
retrainings of the same configuration with new seeds. The second block retrains the same recipe and backbone
and changes only the task interface. The last row queries the model with the demonstrator's own execution
instead of its effect. Published methods read the same held-out demonstration through their own interface
(Appendices~\ref{app:uwm}--\ref{app:zerowam}).}
\label{tab:main_full}
\setlength{\tabcolsep}{3.5pt}
\begin{tabular}{lrrrrl}
\toprule
 & bowl$\to$stove & bottle$\to$cabinet & knob & $\Sigma/90$ & 3 seeds \\
\midrule
\method{}, effect program & 27 & 12 & 1 & \best{40} & 40, 38, 33 \\
\method{}, terminal state only & 30 & 19 & 0 & \best{49} & 49, 50, 32 \\
\midrule
goal image instead of program & 0 & 0 & 0 & 0 & \\
language instead of program & 0 & 0 & 0 & 0 & \\
no program & 0 & 0 & 0 & 0 & \\
\midrule
UWM, demo frames \citep{uwm} & 0 & 0 & 19 & 19 & \\
Instant Policy \citep{instantpolicy} & 0 & 0 & 5 & 5 & \\
ATM, demo tracks \citep{atm} & 0 & 0 & 0 & 0 & \\
ATM, language \citep{atm} & 0 & 0 & 0 & 0 & \\
Zero-WAM, demo video \citep{zerowam} & 0 & 0 & 0 & 0 & \\
\midrule
\method{}, demo execution as query & 28 & 28 & 29 & 85 & \\
\bottomrule
\end{tabular}

\end{table}

\paragraph{Other queries of the same weights.} Given the demonstrator's own end-effector trajectory, the input UWM uses, the same weights complete 85 of 90 (last row of Table~\ref{tab:main_full}): LIBERO episodes of one task start within a median 1.7\,cm of each other, so the demonstrator's motion carries over almost unchanged. The effect program specifies the same tasks with that motion removed.

\paragraph{Replicate trainings.} Retraining with a new seed moves the held-out sum by up to seven episodes for the effect program (40, 38, 33) and by up to 18 for the terminal state (49, 50, 32), and a four-arm bowl sum by up to 22 (Appendix~\ref{app:v100}). Differences are read against the replicate spread of the statistic they concern.

\paragraph{Layout variation.} Over the 110 tasks in the corpus, the manipulated object's starting position differs
between two episodes of a task by a median 2.5\,cm, and in 99\% of tasks it stays under 3\,cm (bowl task: median 1.7\,cm). Layout transfer is measured on Meta-World, where every episode is programmed by another episode's demonstration (\S\ref{sec:beyond}).

\paragraph{Held-out tasks have near neighbours, some nearer than others.} How new a held-out task is depends
on how far its effect lies from the effects seen in training. We take a task's effect to be the median
displacement of the entity that moves. The closest training task to bowl$\to$stove is then 4.2\,cm away,
compared with 25\,cm of bowl travel, and the closest to bottle$\to$cabinet is 10.7\,cm away, compared with 33.
The knob moves 0.3\,cm, and two training tasks have effects that match it to one decimal place. The knob is
also turned in three other LIBERO-90 scenes, where the main checkpoint completes it from its program 10 of 10
times; in the held-out scene it completes 1 of 30.

\paragraph{Success is nearly binary on pick-and-place.} On the bowl and bottle tasks, failures are 400-step
timeouts that leave the object a median 26 and 33\,cm from its programmed destination, with the object still at its start. Successes end 2.1 and 1.6\,cm away, and no episode falls in between. The knob behaves
differently: failures stop a median 4.8\,cm from the target, against 2.0\,cm for a success.

\paragraph{What the seed changes.} To see where the spread between trainings comes from, we compare the three
trainings episode by episode. On the bowl task they agree on all 30 episodes under the terminal-state query
and on 26 of 30 under the full program (24 succeed under every seed, 2 under none). On the bottle they agree on 13: three episodes succeed under all three trainings, ten fail under all three, and seventeen change with the seed. A failure looks the same under every seed, a timeout with the bottle 33 to 34\,cm from its
destination. The seed therefore decides which episodes get the object picked up, and each outcome keeps its form across seeds.

\paragraph{Attributing editability to the corpus.} We compare how closely \method{} follows a commanded
rotation of the contact points with how closely W8 does, the variant that has the contact channel but not the
corpus. The test was fixed before the last two trainings: one further training of each variant (seed 5,
because on two GPUs every lower seed reuses a per-rank seed of an earlier training), 50 episodes per commanded
rotation for all seven trainings on one machine, and an exact one-sided permutation test over four \method{}
and three W8 trainings at $p\le0.05$, which requires every \method{} training to follow more closely than
every W8 training. The two new trainings follow at 0.66 and 0.34; all twelve pairs order the same way, giving
$p{=}1/35{=}0.029$.

\section{Ablation variants}
\label{app:ablations}
\paragraph{Contact channel and paired corpus.} Across the four robot arms, the recipe with both ingredients
has a mean bowl sum of 83.3 over three trainings. The contact channel alone gives 82.0, neither ingredient 78.5, and the corpus alone 63.5 (means of two trainings). The interaction $I$ of prereg 27 is 16.3: the corpus lowers the bowl sum without the contact channel and raises it with the channel. On held-out success the recipe without the corpus completes 51 of 90 and the full recipe 40 (Table~\ref{tab:ablations}); on terminal edits the recipe with neither ingredient follows with slope 0.80 and the corpus-only recipe with 0.81, against 0.71 for \method{} (Appendix~\ref{app:editing}). The two ingredients act on contact editing.

Table~\ref{tab:ablations} collects the variants the main text summarises; the table script reads each entry
from that variant's own evaluation records. Every row is a single training, read against the replicate spread of Appendix~\ref{app:diag}. Variants keep the names used in the project
records. W3 has neither the contact channel nor the paired corpus, W8 has the contact channel only, and W6
the corpus only. v9 uses one contact point, v13 a dense surface sample, and v14 stores contact as affine
weights. R1--R5 retrain the main recipe with one change each, and W0--W5 vary which queries the earlier
recipe trains.

\paragraph{Which queries to train.}
\label{app:queries}
The first block varies which queries are trained, on the earlier backbone's recipe (no contact channel, no
paired corpus). Because it also uses an earlier protocol (re-planning every eight steps, no fixed sampling
seed), it is compared only within itself; the second block re-evaluates three of its variants under the table
protocol. Two differences exceed the spread between trainings. Adding the unconditioned joint query to the
training mix (W5) drops the terminal-state query to 29 of 90, against 56 to 58 without it. Leaving the
terminal-state query out of training (W1) drops the effect program to 22 (21 under the table protocol), against 35 to 46. The zero rows of the main table come from three separately trained variants rather than one
model asked three ways, because the main model never trains the joint query and so cannot be asked to act on
nothing. Given no program, the unconditioned variant still moves the bowl a median 14.5\,cm (a success moves it 24) and ends 36\,cm from the commanded effect: it acts
towards an effect it sampled itself. The goal-image variant moves the bowl 31.8\,cm and the bottle 26.8\,cm
without completing either task.

\paragraph{Contact representation, corpus and forward query.} On the held-out sum, one contact point instead of two (v9, 34), the corpus without the contact channel (W6, 35) and no forward query (W7, 47) stay within seven of the main model's 40, and the contact channel without the corpus (W8) completes 51. On the four-arm bowl sum the main model's 89 is the highest of the five (75, 64, 77 and 74 for v9, W6, W7 and W8). W6, with the corpus alone, completes 35 against 46 for W3 with neither component, with the bottle at 10 of 30 against 18.

\paragraph{Test-time changes.} These changes alter how the trained main model is queried, without retraining
it. Collapsing the two contact points to their midpoint gives four-arm sums of 77 and 40, against 89 and 43.
Feeding each arm its own joint angles, instead of re-expressing proprioception in the training arm's joint
space, keeps the Panda at 27 of 30 but drops the other three arms to 1, 0 and 0. All remaining single
changes were fixed before any of them ran, and only one moves the sum by more than ten: advancing the program window by the demonstration's clock, which gives 27, with the bottle at 1 of 30.

\paragraph{Retrained variants.} Each of these variants retrains the main recipe with one design choice
changed. Training the inverse query alone (R1) completes 41 of 90 and follows the two $\pm45^\circ$ commanded rotations at 0.83 and 0.93, against 0.47 and 0.82 for the main model. Sharing one timestep across all streams (R2) completes 37 and follows at 0.63 and 0.49; removing the execution stream (R3) completes 35, at 0.46 and 0.78; and an effect window of eight steps (R4) completes 42, at 0.47 and 0.92. Writing the effect as displacements from each entity's anchor (R5) reaches the main model's
training loss (0.0615 against 0.0601 at 30k) and completes 4 of 10 episodes of the two drawer training tasks against the main model's 6, and 1 of the 90 held-out episodes; on the bowl task its executions stop above the bowl, so absolute coordinates are the headline choice on LIBERO.

\begin{table}[h]
\centering\small
\caption{\textbf{One program, four arms, no retraining}: successes out of 30 on the bowl / bottle tasks,
split by whether the recipe has the contact channel and the paired multi-solution corpus. Proprioception is
re-expressed in the training arm's joint space, which is the identity on the Panda. Each recipe is trained
twice (seeds 0 and 1) and \method{} three times; the last block averages the sums over trainings.}
\label{tab:arms_full}
\setlength{\tabcolsep}{4pt}
\begin{tabular}{llrrrrrr}
\toprule
ingredients & contact / corpus & Panda & IIWA & Kinova3 & UR5e & $\Sigma$bowl & $\Sigma$bottle \\
\midrule
\multicolumn{8}{l}{\emph{first training}} \\
neither       & --- / ---   & 28/18 & 26/7  & 17/16 & 8/0  & 79 & 41 \\
contact only  & yes / ---   & 27/17 & 21/13 & 24/10 & 2/1  & 74 & 41 \\
corpus only   & --- / yes   & 24/7  & 15/0  & 11/2  & 14/1 & 64 & 10 \\
both (\method{}) & yes / yes   & 27/15 & 24/17 & 22/5  & 16/6 & 89 & 43 \\
\midrule
\multicolumn{8}{l}{\emph{further trainings}} \\
neither       & --- / ---   & 29/18 & 21/13 & 18/9  & 10/1 & 78 & 41 \\
contact only  & yes / ---   & 28/21 & 26/18 & 26/16 & 10/1 & 90 & 56 \\
corpus only   & --- / yes   & 21/12 & 15/1  & 15/1  & 12/0 & 63 & 14 \\
both, second  & yes / yes   & 26/10 & 18/5  & 18/7  & 12/5 & 74 & 27 \\
both, third   & yes / yes   & 26/4  & 18/8  & 17/5  & 11/0 & 72 & 17 \\
\midrule
\multicolumn{8}{l}{\emph{mean over trainings}} \\
neither       & --- / ---   &       &       &       &      & 78.5 & 41.0 \\
contact only  & yes / ---   &       &       &       &      & 82.0 & \best{48.5} \\
corpus only   & --- / yes   &       &       &       &      & 63.5 & 12.0 \\
both (\method{}) & yes / yes   &    &       &       &      & \best{83.3} & 27.7 \\
\bottomrule
\end{tabular}

\end{table}

\begin{table}[h]
\centering\footnotesize
\caption{Ablation variants, one training each unless marked as a second training (seed 1). Effect program:
successes out of 30 on the three held-out tasks and their sum; goal: the terminal-state query's sum over 90;
four arms: sums over Panda, IIWA, Kinova3 and UR5e on the bowl and bottle tasks, as in Table~\ref{tab:arms};
following: mean over the $\pm20^\circ$ and $\pm45^\circ$ commanded rotations, as in Table~\ref{tab:rot};
$^\dagger$: mean over $\pm45^\circ$ only. ``--'': not evaluated under that block's protocol, or (W1, R1) a
query the variant never trains. The first block and the goal column were evaluated earlier, partly on an RTX~A6000; every other entry was evaluated on RTX~3090s. The first block uses a different protocol from the others.}
\label{tab:ablations}
\setlength{\tabcolsep}{3pt}
\begin{tabular}{lrrrrrrrr}
\toprule
 & \multicolumn{4}{c}{effect program} & goal & \multicolumn{2}{c}{four arms} & \\
\cmidrule(lr){2-5}\cmidrule(lr){7-8}
 & bowl & bottle & knob & $\Sigma$ & $\Sigma$ & $\Sigma$bowl & $\Sigma$bottle & following \\
\midrule
\multicolumn{9}{l}{\emph{which queries to train: earlier backbone's recipe; re-plan every 8 steps, no fixed sampling seed}} \\
all six queries (the earlier backbone) & 24 & 12 & 4 & 40 & -- & -- & -- & -- \\
all six, goal share 0.25 (W5) & 16 & 13 & 0 & 29 & 29 & -- & -- & -- \\
inverse, goal, inverse dynamics (W0) & 29 & 6 & 0 & 35 & 58 & -- & -- & -- \\
W0 + partial program (W2) & 24 & 13 & 0 & 37 & 56 & -- & -- & -- \\
W0 + forward (W3) & 26 & 18 & 0 & 44 & 57 & -- & -- & -- \\
inverse and inverse dynamics only (W1) & 10 & 12 & 0 & 22 & -- & -- & -- & -- \\
\midrule \multicolumn{9}{l}{\emph{which queries to train, the same variants under the table protocol}} \\
inverse, goal, inverse dynamics (W0) & 27 & 14 & 0 & 41 & 57 & -- & -- & -- \\
W0 + forward (W3) & 28 & 18 & 0 & 46 & 58 & -- & -- & -- \\
inverse and inverse dynamics only (W1) & 6 & 15 & 0 & 21 & -- & -- & -- & -- \\
\midrule \multicolumn{9}{l}{\emph{contact representation, corpus and forward query: table protocol}} \\
headline: two contact points, paired corpus & 27 & 12 & 1 & 40 & 49 & 89 & 43 & 0.68 \\
one contact point, paired corpus (v9) & 27 & 7 & 0 & 34 & 52 & 75 & 36 & 0.62 \\
two contact points, no paired corpus (W8) & 28 & 19 & 4 & 51 & 57 & 74 & 41 & 0.31 \\
paired corpus, no contact channel (W6) & 25 & 10 & 0 & 35 & 45 & 64 & 10 & -- \\
headline without the forward query (W7) & 29 & 16 & 2 & 47 & 54 & 77 & 32 & -- \\
W0 + forward (W3), second training & 29 & 15 & 0 & 44 & 49 & 78 & 41 & -- \\
no paired corpus (W8), second training & 30 & 19 & 0 & 49 & 53 & 90 & 56 & -- \\
no contact channel (W6), second training & 24 & 6 & 0 & 30 & 44 & 63 & 14 & -- \\
\midrule \multicolumn{9}{l}{\emph{test-time changes to the headline model: table protocol}} \\
contact pair collapsed to its midpoint & -- & -- & -- & -- & -- & 77 & 40 & -- \\
proprioception read natively on each arm & -- & -- & -- & -- & -- & 28 & 16 & -- \\
program window only, no terminal state & 26 & 12 & 0 & 38 & -- & -- & -- & -- \\
clamp only the window's last frame & 27 & 19 & 0 & 46 & -- & -- & -- & -- \\
clamp every fourth frame of the window & 28 & 12 & 0 & 40 & -- & -- & -- & -- \\
window advanced by the demonstration's clock & 25 & 1 & 1 & 27 & -- & -- & -- & -- \\
window never ahead of the scene & 28 & 16 & 1 & 45 & -- & -- & -- & -- \\
re-plan every 8 steps & 26 & 11 & 1 & 38 & -- & -- & -- & -- \\
re-plan every 16 steps & 24 & 13 & 1 & 38 & -- & -- & -- & -- \\
5 sampling steps (of 20) & 27 & 14 & 0 & 41 & -- & -- & -- & -- \\
10 sampling steps & 27 & 13 & 1 & 41 & -- & -- & -- & -- \\
40 sampling steps & 28 & 11 & 1 & 40 & -- & -- & -- & -- \\
best of 8 executions by the forward query & 28 & 12 & 0 & 40 & -- & -- & -- & -- \\
program from another episode of the task & 25 & 9 & 0 & 34 & -- & -- & -- & -- \\
\midrule \multicolumn{9}{l}{\emph{retrained variants of the headline recipe, one training each: table protocol}} \\
inverse query only (R1) & 25 & 16 & 0 & 41 & -- & -- & -- & 0.88$^\dagger$ \\
one timestep shared by all streams (R2) & 26 & 11 & 0 & 37 & 45 & -- & -- & 0.56$^\dagger$ \\
no execution stream (R3) & 25 & 10 & 0 & 35 & 52 & -- & -- & 0.62$^\dagger$ \\
effect window of 8 steps (R4) & 24 & 18 & 0 & 42 & 52 & -- & -- & 0.70$^\dagger$ \\
effect relative to the anchor (R5) & 0 & 1 & 0 & 1 & -- & -- & -- & -- \\
\bottomrule
\end{tabular}

\end{table}
\section{Editing, sampling and selection}
\label{app:editing}
\begin{table}[h]
\centering\small
\caption{Rotating the two contact points about the object axis (task 4, $n{=}30$). Left: successes at
each commanded rotation. Right: the fraction of the commanded rotation that the realised grasp achieved,
measured by replaying each rollout and reading the gripper contacts. The three \method{} rows are the same
configuration trained with three seeds; W8 is trained without the paired corpus.}
\label{tab:rot}
\setlength{\tabcolsep}{3.5pt}
\begin{tabular}{lrrrrrrr@{\hskip 1em}rrrrr}
\toprule
& \multicolumn{7}{c}{successes at commanded rotation} & \multicolumn{5}{c}{realised / commanded} \\
\cmidrule(r){2-8}\cmidrule(l){9-13}
& $0$ & $+20$ & $-20$ & $+45$ & $-45$ & $+90$ & $-90$ & $+20$ & $-20$ & $+45$ & $-45$ & mean \\
\midrule
\method{}, seed 0            & 27 & 22 & 24 & 24 & 24 & 22 & 26 & 0.64 & 0.79 & 0.47 & 0.82 & 0.68 \\
\method{}, seed 1            & 26 & 23 & 21 & 24 & 19 & 24 & 22 & 0.38 & 0.38 & 0.49 & 0.46 & 0.43 \\
\method{}, seed 2            & 27 & 26 & 22 & 27 & 21 & 26 & 16 & 0.73 & 0.51 & 0.63 & 0.49 & 0.59 \\
W8 (contact, no corpus)      & 28 & 26 & 20 & 25 & 11 & 24 & 10 & 0.50 & 0.24 & 0.39 & 0.12 & 0.31 \\
\bottomrule
\end{tabular}

\end{table}

\label{app:select}
\label{app:tables}

\paragraph{Rotation direction.} In Table~\ref{tab:rot} the realised grasp follows rotations in one direction
more closely than in the other. The first training realises 0.64 and 0.47 of the positive rotations but 0.79 and 0.82 of the negative ones. We measured and ruled out five candidate causes of this asymmetry
(Appendix~\ref{app:prereg}). The second training shows none and the third reverses it, so the direction
of the asymmetry is set by the training seed.

\begin{figure}[h]
\centering
\includegraphics[width=\textwidth]{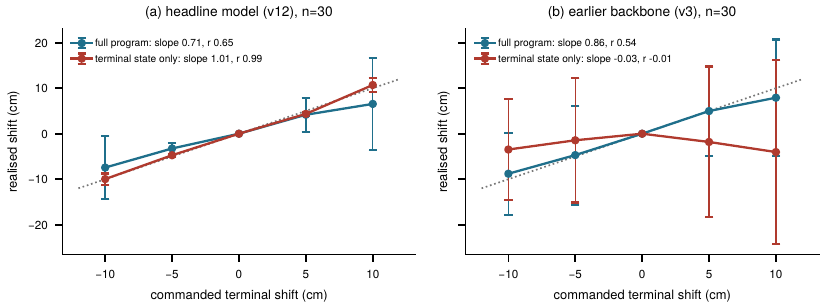}
\caption{Terminal shift under the two queries, bowl task, 30 episodes: mean and standard deviation of the
realised shift, measured as in Figure~\ref{fig:edit}a. The slope is the least-squares fit through the
origin over the shifted settings; $r$ is the correlation over all 150 episode--shift pairs. (a) The headline
model. (b) The earlier backbone, v3.}
\label{fig:terminal}

\end{figure}

\paragraph{Terminal edits under the terminal-only query.} Shifting the terminal state $\Gstar$ should move
where the object is placed, and it does so even when the terminal state is the whole program (the
terminal-only query). Over 30 episodes shifted by $\pm5$ and $\pm10$\,cm, the realised shift follows the
commanded one with slope 1.01 ($r{=}0.99$) under the terminal-only query and with slope 0.71 ($r{=}0.65$)
under the full program (Figure~\ref{fig:terminal}). At $\pm10$\,cm the terminal-only query puts the bowl exactly where it was sent, beside the stove, where the benchmark's predicate fires in 1 and 5 of 30 episodes. The earlier backbone (v3, evaluated under its own 300-step protocol) follows terminal edits through its full program, with slope 0.86 (slope $-0.03$ under the terminal-only query). Waypoints and contact points are written into the effect window, which the terminal state does not contain.

\paragraph{Terminal edits across ingredients.} We next asked whether following terminal edits depends on
either training ingredient, the contact channel or the paired corpus. It depends on neither
(Table~\ref{tab:terminal_arms}). Given the full program, the variant trained with neither ingredient follows
the shift with slope 0.80 and the variant trained with the corpus only with slope 0.81, compared with 0.71
for \method{}; given the terminal state alone, the same three reach 0.95, 0.90 and 1.01. The two ingredients
matter instead for how closely contact edits are followed.

\begin{table}[h]
\centering\footnotesize
\caption{Terminal edits for the training variants of Table~\ref{tab:arms} (bowl task, first training of each
variant, protocol of Table~\ref{tab:main}): successes of 30 at each commanded shift of $\Gstar$, and the slope
of realised against commanded shift (fitted through the origin, with each episode paired against its
unshifted run).}
\label{tab:terminal_arms}
\setlength{\tabcolsep}{4pt}
\begin{tabular}{llrrrrrrr}
\toprule
recipe & query & $-10$ & $-5$ & 0 & $+5$ & $+10$\,cm & slope & $r$ \\
\midrule
\method{} & program & 8 & 27 & 27 & 28 & 22 & 0.71 & 0.65 \\
 & terminal & 1 & 30 & 30 & 30 & 5 & 1.01 & 0.99 \\
neither ingredient (W3) & program & 10 & 27 & 28 & 27 & 20 & 0.80 & 0.89 \\
 & terminal & 3 & 29 & 30 & 30 & 16 & 0.95 & 0.96 \\
corpus only (W6) & program & 6 & 26 & 25 & 27 & 24 & 0.81 & 0.67 \\
 & terminal & 5 & 30 & 30 & 30 & 18 & 0.90 & 0.99 \\
\bottomrule
\end{tabular}

\end{table}

\paragraph{A reversed program.} Running a program backwards asks for the opposite effect. Started from the
demonstration's initial state, the reversed program asks the bowl to stay where it is
(Figure~\ref{fig:reverse}), so we place the bowl at the program's end before the episode and ask the reversed
program to carry it back (Table~\ref{tab:reverse_end}): 2 of 30 episodes reach the start, against 1 of 30
with no program, with the bowl a median 25.7\,cm from it. Programs are edited through their terminal state
and contact points.

\begin{table}[h]
\centering\footnotesize
\caption{The bowl is placed at the program's end (on the stove, within 2.1\,mm) before the episode starts.
An episode succeeds if the bowl's centre ends within 5\,cm of the target; 30 episodes, protocol of
Table~\ref{tab:main}.}
\label{tab:reverse_end}
\setlength{\tabcolsep}{4pt}
\begin{tabular}{lrrr}
\toprule
 & success & median final distance & bowl raised $>3$\,cm \\
\midrule
reversed program (target: the demonstration's start) & 2 & 25.7\,cm & 4 \\
no program (target: the demonstration's start) & 1 & 24.8\,cm & 3 \\
forward program (target: the stove; already there) & 30 & 0.0\,cm & -- \\
\bottomrule
\end{tabular}

\end{table}

\paragraph{How much program error the interface absorbs.}
\label{app:noise}
This experiment measures how success changes when the programmed destination is wrong. We added isotropic
Gaussian noise of standard deviation $\sigma$ to the destination, using the same 30 bowl episodes at every
noise level and the protocol of the main table. At $\sigma{=}0$ the true program is written to a file and
clamped through the same path as the noisy ones; it completes 27 of 30, as in the table. At $\sigma{=}1$\,cm success is 28, and at 2 and 4\,cm it is 24 and 20: flat up to a centimetre and degrading beyond two.

\paragraph{Live keypoints from rendered RGB-D.}
\label{app:live}
Everywhere else in the closed loop, the live keypoints in $O_t$ are read from simulator state. Three parts of
the loop consume them: the anchor, the entities the pointer matches against the program, and the grasp gate.
Here we segment them from rendered images as well, to measure what perception costs. At every environment
step, SAM~2 tracks each entity the program moves over a two-frame window (the previous and the current
frame), prompted with the previous mask. The mask is back-projected with the rendered depth, and the entity's
centre is the mid-point of the 5th to 95th percentiles of the resulting points. We follow the conventions of
the program extractor of \S\ref{sec:real}: the first-frame prompt is each entity's reset-time centre
projected into the image, in place of a click; box size and orientation keep their reset values and only the
centre is updated; entities the program does not move keep their reset pose. The bowl task uses the front
camera and the bottle task the front and wrist cameras, at $256^2$; perception takes about 0.2\,s per step.
There are three settings: P0 reads both the program and the live keypoints from state, P1 segments the live
keypoints, and P2 also extracts the program from pixels. All three share the checkpoint, the 30 episodes per
task and the protocol of Table~\ref{tab:main}. The bowl programs of P2 are the single-view programs of
\S\ref{sec:real}; the bottle programs use both views.

Segmenting only the live keypoints gives 16 bowl and 9 bottle successes, against 27 and 12 from state; with the program also from pixels, the counts are 25 and 15 (Table~\ref{tab:live}). On the bottle the tracked centre is off by a median 3.2 (P1) and 3.4\,cm (P2), the mask is lost for at least one step in 1 (P1) and 2 (P2) episodes, and successful episodes track to a median 2.1 and 3.0\,cm against 4.6 and 5.9\,cm for failed ones. On the bowl the median error stays below 1\,cm, but the per-episode maximum exceeds 5\,cm in 14 (P1) and 11 (P2) of the 30 episodes.

On the bowl, the pointer decides most P1 episodes. To locate it, we re-ran P1 and sent the
segmented keypoints to one consumer at a time. When only the pointer reads simulator state, 26 of 30 episodes
succeed; when only the pointer reads the segmented keypoints, 18 succeed; when every consumer reads them, 16
succeed. In 6 of the 14 failures of that re-run, the pointer stops between program frames 34 and 45, just
before the programmed lift, and the bowl rises at most 0.4\,cm. The pointer advances past the program's
first motion only when the matched effect does, and while it waits the segmented centre sits 1.0 to 1.9\,cm
from the true one, almost all of it sideways (medians 1.4\,cm horizontal, 0.2\,cm vertical), which keeps the
match on the frames before the lift. In 3 other failures the mask drifts off a bowl that has not moved, and
the segmented centre rises by 11 to 12\,cm.

\begin{table}[h]
\centering\small
\caption{Live keypoints from rendered RGB-D (pre-registration 28): successes of 30, and the error of the live
centre of the manipulated entity against simulator state. Error: median over episodes of the per-episode
median and of the per-episode maximum. Lost: episodes in which the mask vanished for at least one step.}
\label{tab:live}
\setlength{\tabcolsep}{4pt}
\begin{tabular}{llrrrrrr}
\toprule
 & & & & & \multicolumn{3}{c}{live error, bowl / bottle} \\
\cmidrule(lr){6-8}
cell & program / live keypoints & bowl & bottle & $\Sigma$ & median (cm) & max (cm) & lost \\
\midrule
P0 & state / state   & 27 & 12 & 39 & -- & -- & -- \\
P1 & state / pixels  & 16 & 9  & 25 & 0.9 / 3.2 & 4.0 / 13.2 & 0 / 1 \\
P2 & pixels / pixels & 25 & 15 & 40 & 0.6 / 3.4 & 2.3 / 10.7 & 1 / 2 \\
\bottomrule
\end{tabular}

\end{table}

\paragraph{Ranking sampled executions.} Since the model samples executions, one can draw several and keep the
one whose forward-predicted effect best matches the program. Ranking eight draws this way completes 44 of 90
episodes, and the first draw 47. Over 6{,}142 re-planning steps each of the eight draws is chosen 716 to 851
times, close to the 768 of a uniform choice: the draws are interchangeable along the path being scored, and
the diversity described in \S\ref{sec:cross} lies in the grasp.

\paragraph{Re-planning interval.} We chose the re-planning interval on the backbone that preceded the headline
model. At a common 400-step cap, re-solving every 4, 8 or 16 steps completes 44, 40 and 31 of 90 episodes.
Four steps is best on the total (44 against 40 and 31) and on the bottle (20 against 12 and 6), so every table re-solves every four steps.
On the headline model, intervals of 8 and 16 give 38 and 38.

\section{Disturbance}
\label{app:disturb}
Table~\ref{tab:disturb} lists every disturbance setting evaluated on the headline model. The push is
requested at step 25 and applied at the next re-planning boundary: step 28 when re-planning every four steps,
step 32 when re-planning every eight or sixteen. The replay baseline executes the recorded actions of the
episode's own demonstration open loop.

\begin{table}[h]
\centering\footnotesize
\caption{A 5\,cm push on the headline model: successes of 30, undisturbed and pushed. The push lands at
step 28 unless noted otherwise. Variant G2 was designed after the results of the first gate had been seen.
Effect-program rows with the gate were evaluated on RTX~3090s, the other rows on an RTX~A6000.}
\label{tab:disturb}
\setlength{\tabcolsep}{4pt}
\begin{tabular}{lrrrr}
\toprule
 & \multicolumn{2}{c}{bowl$\to$stove} & \multicolumn{2}{c}{bottle$\to$top} \\
\cmidrule(lr){2-3}\cmidrule(lr){4-5}
 & undisturbed & pushed & undisturbed & pushed \\
\midrule
replay of the demonstration's actions & 29 & 6 & 28 & 0 \\
effect program & 27 & 9 & 12 & 14 \\
\quad re-plan every 8 steps (push at step 32) & -- & 18 & -- & -- \\
\quad re-plan every 16 steps (push at step 32) & -- & 21 & -- & -- \\
\quad G2: no nearest-neighbour matching before engagement & 28 & 12 & 14 & 17 \\
effect program without the engagement gate, anchored once & 26 & 3 & 13 & 4 \\
\quad re-anchored at every re-plan & 26 & 10 & 13 & 7 \\
\midrule
\multicolumn{5}{l}{\emph{push at step 32, re-plan every 4 steps}} \\
replay of the demonstration's actions & 29 & 8 & 28 & 1 \\
effect program & 27 & 12 & 12 & 16 \\
effect program without the engagement gate, anchored once & 26 & 6 & 13 & 4 \\
\bottomrule
\end{tabular}

\end{table}

\begin{figure}[h]
\centering
\includegraphics[width=\textwidth]{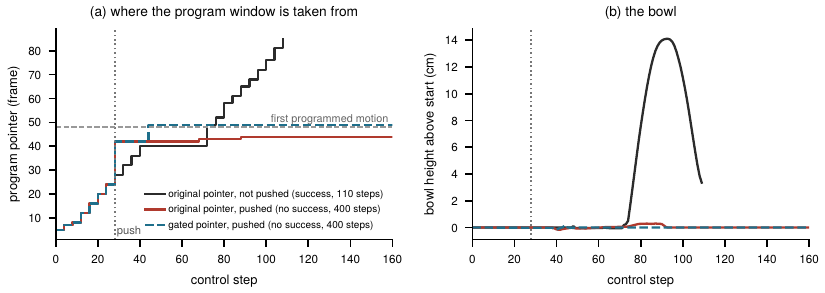}
\caption{Episode 201 of the bowl task, pushed 5\,cm at step 28 (dotted). (a) The program frame selected
at every re-plan. Undisturbed, the pointer without the gate passes the program's first motion (dashed, frame
48) and the bowl is lifted. Pushed, the pointer jumps from frame 24 to 42 both with and without the gate; with the gate it moves to the program's engagement frame 49 at step 44, and without it stays near frame 42. (b) The bowl's height above its start: when pushed, it is never lifted.}
\label{fig:pointer}

\end{figure}

\paragraph{What the gate prevents.} To see why a push defeats the pointer without the gate, we re-ran without
the gate four bowl episodes that succeed undisturbed and fail when pushed (201, 204, 205, 207), recording the
program frame the pointer selects at every re-plan. In all four, the pointer jumps at the push from frame 24
to frames 40--49 and stays there, and the bowl is never lifted (Figure~\ref{fig:pointer}). In episode 205 the gripper is still 26\,cm from the bowl when the push lands, so the jump comes from the matching rather than from the grasp.
After the first object motion the pointer matches the live object against the program, and a push moves the
object too. The programmed frame in which the bowl has started along its route then becomes the nearest
match, and the push is read as progress. The gate of \S\ref{sec:infer} blocks that advance until the object
is held.

\paragraph{A stricter gate.} Figure~\ref{fig:pointer} also shows that in episode 201 the pointer jumps even
with the gate, from frame 24 to 42, stopping just short of the program's first motion (frame 48), which the gate blocks; at engagement it moves to frame 49, but the bowl is not lifted. A stricter variant, G2, therefore uses no nearest-neighbour matching at all before engagement.
Pushed, it completes 29 of 60 episodes. Undisturbed, it completes 43, but it changes seven outcomes relative
to the main table, more than the five the criterion allows, so \S\ref{sec:infer} uses the first gate.

\paragraph{Interval and the earlier backbone.} Re-planning every 8 or 16 steps completes 18 and 21 of 30
pushed bowl episodes, against 9 when re-planning every four. At those intervals the push lands at step 32, so these rows combine the interval with the later push. On the earlier backbone,
the comparison variant was the model's inverse-dynamics query, given the demonstrator's end-effector
trajectory advanced by the clock. It completed 56 of 60 undisturbed episodes and, pushed, 13 and 3 of 30 on
the bowl and the bottle, against 20 and 8 for that backbone's effect program.

\section{Published baselines on the held-out protocol}
\label{app:baselines}
\begin{figure}[!tbp]
\centering
\includegraphics[width=0.79\textwidth]{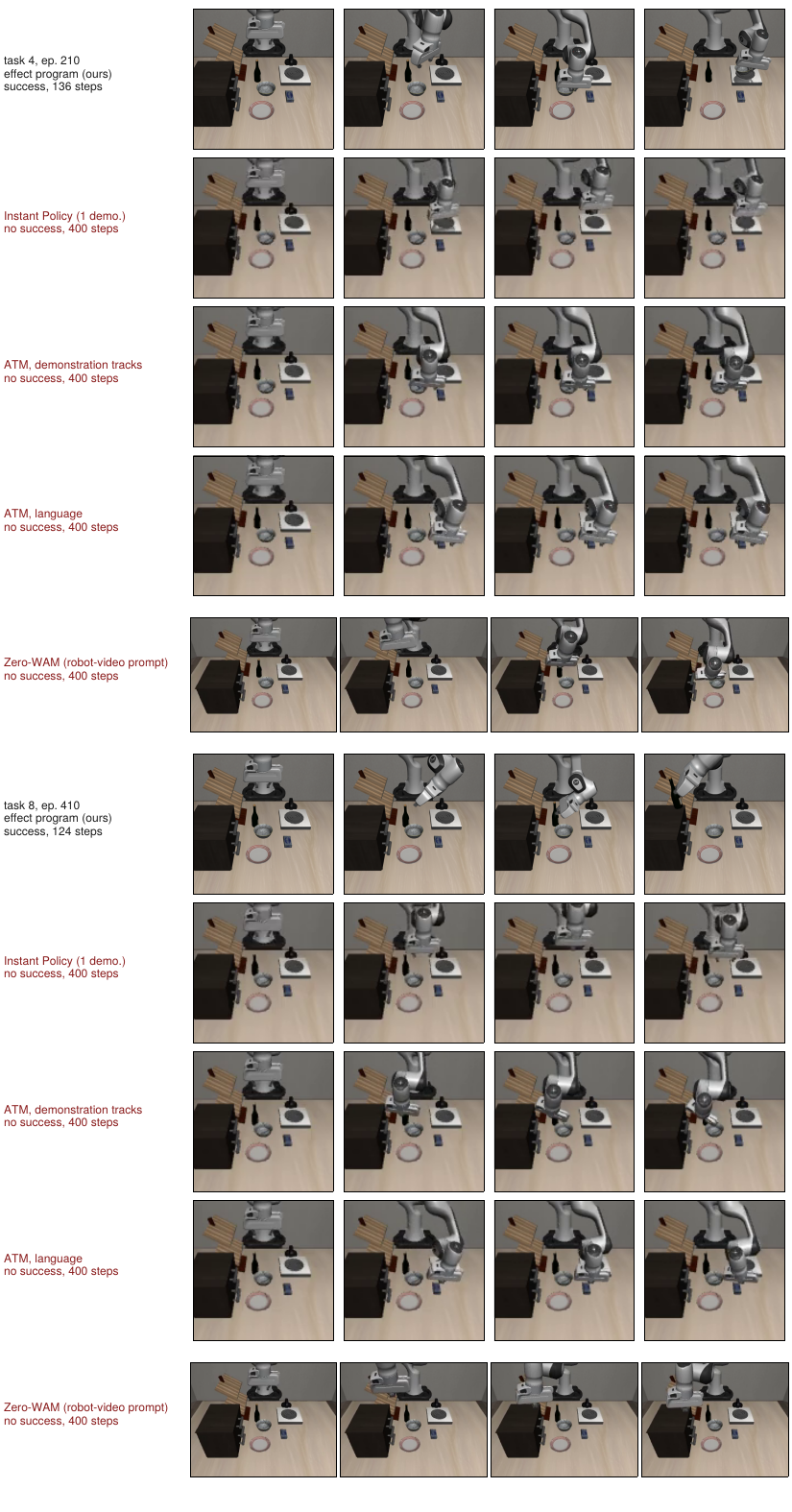}
\caption{Two episodes (bowl episode 210, bottle episode 410) as each demonstration-reading method
executes them; every row shows the outcome recorded in that method's own evaluation. The episodes were picked
by the selection rule of Appendix~\ref{app:gallery}, which no knob episode satisfies. UWM has no row because
it was recorded only for episode 200.}
\label{fig:baselines}

\end{figure}
This appendix describes how each published method was run on our held-out protocol. All baselines use the same episodes and the same success test: the first 30
held-out demonstrations of each task, each episode reset to its demonstration's initial state, a limit of
400 control steps, and the BDDL success check.

\subsection{UWM and the demonstrator's own trajectory}
\label{app:uwm}
UWM can be conditioned on future frames, so it reads a demonstration directly. We query the released
checkpoint through its own \texttt{sample\_inverse\_dynamics}, giving it the demonstration's future frames as
the target and advancing that target as the execution proceeds; we run it both at its own chunk length and at
ours. Two controls show what its row means. The first checks our evaluation loop: on two tasks UWM was
trained on, it succeeds 18 of 20 times. The second concerns what UWM has seen. Its
pretraining corpus, LIBERO-90, contains two tasks whose effect is essentially the knob's, and the knob is
the one task where it does well, with 19 of 30 (12 at our chunk length). It saw neither LIBERO-10 nor our
seven LIBERO-Goal training tasks, which is why the other tasks stay out of reach. For reference, under the
same setup our headline weights complete 28, 28 and 29 of 30, and replaying the demonstration's recorded
actions completes 29, 28 and 30.

If the demonstrator is the same robot, the simplest way to use a demonstration is to copy its trajectory,
so we also measure how well that works. We use the earlier backbone with re-planning every eight steps and
take the demonstration from a \emph{different} episode of the same task. On the bowl and the bottle, the
source episode's end-effector trajectory scores 27 and 16, its recorded actions score 26 and 23, and the
trajectory translated to the new object's position scores 30 and 24; our effect program scores 21 and 12.
Copying works this well because two LIBERO episodes of the same task differ by a median of only 1.7\,cm, so a translated trajectory is already close to correct.

\subsection{Instant Policy}
\label{app:ip}
We run the released checkpoint unchanged and adapt only its input and output to LIBERO.
\emph{Input}: Instant Policy reads point clouds, which we build from the depth and element segmentation of
three static LIBERO cameras. Following the convention of its RLBench evaluation, we keep every geom that
belongs to neither the robot nor the table, back-project it to the world frame, and crop it to the table's
footprint and the space above it. The resulting cloud of each object lies a median 4.4\,cm from the object's
centre in the simulator. The demonstration given in context is the episode's own, replayed state by state
and observed with the same cameras. \emph{Output}: each predicted end-effector pose is reached through
LIBERO's operational-space controller in one to four control steps, with a median residual of 4\,mm and
0.2$^\circ$. Because these steps count against the same 400-step limit, an episode allows only about fifteen
queries. Tripling the step budget and giving two demonstrations
yields 0, 0 and 8 of 30, against 0, 0 and 7 at 400 steps. 

\subsection{ATM}
\label{app:atm}
No LIBERO checkpoint of ATM covers our tasks, so we train it with its released code on the 90 LIBERO-90
tasks and the seven LIBERO-Goal training tasks, with 45 training and 5 validation demonstrations each.
LIBERO-10 is left out because CoTracker keeps the whole demonstration in memory, and LIBERO-10's
300--460-frame demonstrations need 17--27\,GiB, more than our 24\,GiB cards hold.  ATM has two stages. The track transformer keeps ATM's
configuration and per-GPU batch on three GPUs, for a global batch of 1{,}536 instead of 2{,}048. The
track-guided policy runs out of memory at the recipe's batch of 128 and at 64 per GPU, so it trains at 32 per
GPU on two GPUs. Each stage ran under a wall-clock limit set in advance, 36 and 24\,h, and we use the
checkpoint that its own validation loss selects. The track transformer reached epoch 40 of 144 and the policy
epoch 4 of 5; measured in samples, one policy epoch here equals about 4.5 epochs of the recipe. Before any
number was produced we fixed one bug in the released track dataset: its translation augmentation shifted the
cached tracks in place, so each sample inherited the shifts of earlier ones. We now copy the slice before
augmenting it.

We evaluate two variants. \emph{Language} is ATM as published. \emph{Demonstration tracks} keeps the
language but replaces the predicted tracks at every step with the CoTracker tracks of the episode's own
demonstration (32 points, taken at the demonstration frame aligned with the current step), so that ATM reads
the demonstration the way our model does. Neither variant completes a held-out episode. To confirm that the
trained checkpoint acts at all, we set a threshold before seeing its numbers: at least 7 of 70 successes on
training tasks. It passes, completing 12 of 70 training-task episodes from language and 3 of 70 with
demonstration tracks, so the held-out failures come from the new tasks rather than from a broken checkpoint.

\subsection{Zero-WAM}
\label{app:zerowam}
Zero-WAM released two checkpoints of about 11B parameters, one pretrained and one post-trained on RoboTwin's
bimanual robot, and no single-arm or LIBERO checkpoint. Retraining needs about 173\,GB per GPU, so we run a
released checkpoint zero-shot. The prompt is the episode's own demonstration from the agentview camera at 12
frames per second, with an empty language prompt. Agentview and eye-in-hand fill its high and left-wrist
views, and the right-wrist view is set to zeros. We execute its left-arm end-effector target and gripper
command through the pose servo of the Instant Policy port; a fixed transform, written before any rollout,
rotates RoboTwin's world frame by $90^\circ$ into LIBERO's. Three checks were fixed in advance to make sure
this pipeline works: the checkpoint answers a dummy query; replaying each demonstration's own end-effector
trajectory through the decoding path and the servo succeeds in 9 of 9 episodes; and smoke episodes produce
bounded actions. We chose between the two checkpoints by a rule also set in advance, the median cosine
between the first eight predicted displacements and the demonstration's next 3\,cm of motion; the
post-trained checkpoint won, 0.599 against 0.555. It completes 0 of 30 on each task, with median progress 0.
The arm does move, ending a median 23.7, 18.9 and 27.6\,cm from its start, but no target object moves more
than 0.06\,cm. The setting is far from the checkpoint's training domain: the LIBERO simulator, the single
Panda and its cameras are all new to it, and the prompt is a robot video rather than a human one.

\section{Further LIBERO splits}
\label{app:fold2}
\paragraph{A second LIBERO-Goal split.} A second split holds out three other tasks: before any training, \texttt{numpy.random.default\_rng(2027)} picked
three of the seven training tasks to hold out instead. They are: open the middle drawer; open the top drawer
and put the bowl inside; and put the bowl on the plate. The recipe is the headline one with the held-out tasks
exchanged. Because the bowl-on-plate task is now held out, the paired corpus built on it is replaced by one
built on tasks 4, 7 and 8 (Table~\ref{tab:fold2}). The program completes the bowl-on-plate task in 27 of 30 episodes. The two drawer tasks are articulated effects of the kind described in Appendix~\ref{app:diag}; in the top-drawer task the object moves a median 9.0 and 14.5\,cm under the two queries.

\begin{table}[h]
\centering\small
\caption{The second LIBERO-Goal split, with held-out tasks drawn at random: successes out of 30, one
training run, table protocol.}
\label{tab:fold2}
\begin{tabular}{lrrrr}
\toprule
query & middle drawer & top drawer, bowl inside & bowl$\to$plate & $\Sigma/90$ \\
\midrule
effect program & 0 & 0 & 27 & 27 \\
terminal state only & 0 & 0 & 29 & 29 \\
\bottomrule
\end{tabular}

\end{table}

\paragraph{LIBERO-Object.}
\label{app:object}
This split tests whether one skill transfers to objects the weights have never manipulated. Before any
number was read, we held out three objects (a bottle, a box and a carton), and trained the headline recipe on
the rest of the suite together with the LIBERO corpus.

\begin{table}[h]
\centering\footnotesize
\caption{The second held-out split: one skill, three objects the weights never manipulated, 30 episodes
each. The upper model was trained twice with the same recipe (seeds 0 and 1). Effect-program rows were evaluated on RTX~3090s and terminal-state rows on V100s. The lower pair never saw LIBERO-Object:
across its 90 episodes the object's median displacement is 0.0\,cm, so the object stays where it began, a
median 17\,cm from the target, against 5\,cm for the upper pair.}
\label{tab:object}
\setlength{\tabcolsep}{4pt}
\begin{tabular}{llcrrrr}
\toprule
model & query & training & bbq sauce & chocolate & milk & $\Sigma/90$ \\
 & & & (bottle) & pudding (box) & (carton) & \\
\midrule
\multirow{4}{*}{\method{} + LIBERO-Object} & effect program & 1 & 20 & 29 & 18 & \best{67} \\ & effect program & 2 & 19 & 28 & 18 & \best{65} \\
 & terminal state only & 1 & 29 & 30 & 30 & \best{89} \\
 & terminal state only & 2 & 25 & 30 & 29 & \best{84} \\
\midrule
\multirow{2}{*}{\method{}, no object data} & effect program & 1 & 0 & 0 & 0 & 0 \\
 & terminal state only & 1 & 0 & 0 & 0 & 0 \\
\bottomrule
\end{tabular}

\end{table}

\section{Meta-World}
\label{app:metaworld}
Meta-World defines its own splits. ML45 trains on 45 classes and tests on bin-picking, box-close,
hand-insert, door-lock and door-unlock; ML10 trains on 10 and tests on door-close, drawer-open, lever-pull,
shelf-place and sweep-into. The robot is a Sawyer with a four-dimensional action in $[-1,1]$, and the
demonstrations come from the benchmark's scripted experts.

\paragraph{Corpus, entities and checks.} The training corpus takes the 50 tasks listed for each class and
records each one until \texttt{info['success']} fires, plus ten further steps (ML45: 2{,}462 episodes,
211{,}816 steps, 36 to 50 episodes per training class). The program tracks every body other than the robot
that moves on a free, hinge or slide joint, and describes each with the same nine keypoints. Because
box-close's scripted expert fails on 6 of its 50 tasks, every class is evaluated on the first 30 tasks its
expert solved; this list was fixed from the experts' record before any model was evaluated. Two checks
confirm that evaluation reproduces the recorded episodes: resetting from the stored \texttt{rand\_vec}
reproduces every first frame to within $10^{-4}$\,m, and replaying the recorded actions reaches success at
the recorded step in all 150 sampled episodes. Meta-World also draws its own target marker in the image.
Every variant sees it, including the reference, which receives no task specification.

\paragraph{Evaluation.} The pointer rule is chosen on validation episodes of the training classes (time
pointer: 82 of 91; stage pointer: 69). Each test class is run in two layouts. In the \emph{own layout}, an
episode is programmed by its own demonstration. In the \emph{transferred layout}, episode $j$ is programmed
by the demonstration of episode $j{+}1$ of the same class; at every re-plan the program is translated to the
object's current position, with the offset decaying to zero at $\Gstar$. All runs use $K_{\text{exec}}{=}4$,
a deterministic seed and 500 steps. The published rows reach the test class differently: they fine-tune for
200 gradient steps on ten expert episodes of the class, observe a point cloud, and report the mean of their
five best checkpoints over three seeds. Our model changes no weights, reads one demonstration, and takes
entities from simulator state.

\begin{table}[h]
\centering\footnotesize
\caption{Meta-World ML10, five test classes, success rate in per cent. Published rows are as reported by
\citet{dami}. Our rows use 30 episodes per class, one training run and a checkpoint fixed before evaluation;
the reference is the same recipe trained on the observation alone. Bold marks a result above every published
row; the relative rows are those reported in \S\ref{sec:beyond}. The \emph{seeds 0--2} rows add two more
trainings of that recipe (seeds 1 and 2, trained after the row above them had been read) and give the mean
per cent and the lowest and highest count out of 30.}
\label{tab:metaworld_ml10}
\setlength{\tabcolsep}{3pt}
\begin{tabular}{lrrrrrr}
\toprule
 & door-close & drawer-open & lever-pull & shelf-place & sweep-into & mean \\
\midrule
Meta-World's own \citep{metaworld} & 94.0 & 22.7 & 5.3 & 0.0 & 24.7 & 29.3 \\
DP3 & 100.0 & 70.3 & 44.7 & 12.7 & 16.7 & 48.9 \\
Mamba & 100.0 & 32.3 & 27.7 & 10.0 & 9.7 & 35.9 \\
FreqPolicy & 93.0 & 98.0 & 18.7 & 7.0 & 15.7 & 46.5 \\
FlowPolicy & 100.0 & 13.3 & 17.7 & 0.0 & 6.7 & 27.5 \\
DAMI \citep{dami} & 100.0 & 100.0 & 45.3 & 10.7 & 28.0 & 56.8 \\
effect program, own layout & 100.0 & 0.0 & 0.0 & \textbf{83.3} & \textbf{100.0} & 56.7 \\
effect program, transferred layout & 100.0 & 0.0 & 0.0 & \textbf{50.0} & \textbf{93.3} & 48.7 \\
relative (R5), own layout & 100.0 & 53.3 & 0.0 & \textbf{53.3} & \textbf{100.0} & 61.3 \\
\quad seeds 0--2, mean & 100.0 & 61.1 & 0.0 & \textbf{45.6} & \textbf{100.0} & 61.3 \\
\quad seeds 0--2, lowest--highest of 30 & 30--30 & 9--30 & 0--0 & 11--16 & 30--30 & \\
relative (R5), transferred layout & 100.0 & 33.3 & 3.3 & \textbf{36.7} & \textbf{100.0} & 54.7 \\
\quad seeds 0--2, mean & 76.7 & 47.8 & 2.2 & \textbf{22.2} & \textbf{100.0} & 49.8 \\
\quad seeds 0--2, lowest--highest of 30 & 17--30 & 8--25 & 0--1 & 3--11 & 30--30 & \\
joint state, own layout & 93.3 & 60.0 & 0.0 & \textbf{46.7} & \textbf{96.7} & 59.3 \\
joint state, transferred layout & 53.3 & 0.0 & 0.0 & \textbf{20.0} & \textbf{96.7} & 34.0 \\
reference, observation only & 0.0 & 23.3 & 3.3 & 0.0 & 0.0 & 5.3 \\
\bottomrule
\end{tabular}
\end{table}

\paragraph{ML10.} Table~\ref{tab:metaworld_ml10} gives every ML10 row. Under the transfer the absolute-coordinate variant completes shelf-place at 50.0\%, sweep-into
at 93.3\% and door-close at 100\%. Drawer-open and lever-pull pull a handle along its joint, a motion that none
of the ten ML10 training classes contains; the ML45 variant, whose training classes include both, completes
them 30 of 30 from its program. Writing the effect relative to the anchor opens the drawer on ML10 as well (16
of 30 in the own layout, 10 under the transfer), and this relative variant is the one the reporting rule
selects (82 of 150, against 73 and 51).

\paragraph{The reference given the program.} Because the reference is trained with no stream clamped, it
can still be handed the effect program at test time. This separates what the program contributes from what
training with the clamped stream contributes. With the program, the reference completes sweep-into 27 of 30
in the own layout and 28 under the transfer. The reported variant completes 30 and 30, and the same
reference weights without the program complete 0 and 0, so on sweep-into the program alone carries the
result. On shelf-place the reference completes 2 and 2 against 16 and 11; there, training with the stream
clamped accounts for at least 14 and 9 of 30. This comparison was added after the sweep-into numbers had been seen (Appendix~\ref{app:prereg}).

\begin{table}[h]
\centering\footnotesize
\caption{Meta-World ML45, five test classes, success rate in per cent. The upper block reproduces
Table~III of \citet{dami}: its own method and the five baselines it runs, including the Meta-World RL baseline, at 20 episodes per class on three
seeds, averaging the five best checkpoints after 200 adaptation steps on the class. The lower block is this
paper, at 30 episodes on one seed and one checkpoint, with no weight change at test time; the
\emph{transferred layout} programs each episode with another episode's demonstration. The \emph{seeds 0--2}
rows add two more trainings of the joint-state recipe (seeds 1 and 2, trained after the row above them had
been read) and give the mean per cent and the lowest and highest count out of 30. Bold marks a result above
every published row.}
\label{tab:metaworld_pub}
\setlength{\tabcolsep}{3pt}
\begin{tabular}{lrrrrrr}
\toprule
 & bin-picking & box-close & hand-insert & door-lock & door-unlock & mean \\
\midrule
Meta-World's own \citep{metaworld} & 0.3 & 1.7 & 10.3 & 42.7 & 7.7 & 12.5 \\
DP3 & 9.3 & 26.7 & 6.7 & 51.0 & 94.0 & 37.5 \\
Mamba & 9.7 & 12.7 & 5.3 & 22.7 & 34.7 & 17.0 \\
FreqPolicy & 11.0 & 25.3 & 13.7 & 11.3 & 6.0 & 13.5 \\
FlowPolicy & 9.0 & 13.3 & 6.0 & 15.0 & 22.0 & 13.1 \\
DAMI \citep{dami} & 13.7 & 22.3 & 13.7 & 39.3 & 98.7 & 37.5 \\
\midrule
effect program, own layout & \textbf{50.0} & 6.7 & \textbf{60.0} & 36.7 & 6.7 & 32.0 \\
effect program, transferred layout & \textbf{50.0} & 6.7 & \textbf{36.7} & 20.0 & 20.0 & 26.7 \\
joint state, own layout & \textbf{73.3} & \textbf{36.7} & \textbf{53.3} & 40.0 & 3.3 & 41.3 \\
\quad seeds 0--2, mean & \textbf{64.4} & \textbf{44.4} & \textbf{62.2} & 23.3 & 3.3 & 39.5 \\
\quad seeds 0--2, lowest--highest of 30 & 16--22 & 5--24 & 16--20 & 3--12 & 0--2 & \\
joint state, transferred layout & \textbf{76.7} & 10.0 & \textbf{36.7} & 20.0 & 26.7 & 34.0 \\
\quad seeds 0--2, mean & \textbf{52.2} & \textbf{33.3} & \textbf{38.9} & 17.8 & 27.8 & 34.0 \\
\quad seeds 0--2, lowest--highest of 30 & 9--23 & 3--18 & 11--13 & 3--7 & 6--11 & \\
reference, observation only & 0.0 & 43.3 & 30.0 & 70.0 & 13.3 & 31.3 \\
\bottomrule
\end{tabular}
\end{table}

\paragraph{ML45.} Table~\ref{tab:metaworld_pub} compares the ML45 rows with the published ones, and Table~\ref{tab:metaworld} the two recipes and the reference. In the joint-state recipe, the effect and terminal tokens of an entity on a joint also carry
that joint's position. Across the five classes this recipe has the higher sum: 41.3 against 32.0\% in the own
layout and 34.0 against 26.7 under the transfer. Bin-picking passes the three-part rule, and hand-insert is
above every published row at 36.7. On box-close and door-lock the reference, which receives no task
specification, is already above every published row (43.3 and 70.0 against 26.7 and 51.0), so these two
classes reflect a scene prior. Door-unlock is a lever effect; given the demonstrator's execution, the absolute
variant completes it 26 of 30.

\begin{table}[h]
\centering\footnotesize
\caption{Meta-World ML45, five test classes, 30 episodes each, scored by the benchmark's own success
check. Both models share one recipe and schedule, are trained on the 45 training classes and are compared
only with each other; the third row is that recipe trained on the observation alone.}
\label{tab:metaworld}
\setlength{\tabcolsep}{4pt}
\begin{tabular}{lrrrrrr}
\toprule
 & bin-picking & box-close & hand-insert & door-lock & door-unlock & $\Sigma/150$ \\
\midrule
effect program & 15 & 2 & 18 & 11 & 2 & 48 \\
terminal state only & 0 & 13 & 16 & 4 & 6 & 39 \\
reference, observation only & 0 & 13 & 9 & 21 & 4 & 47 \\
joint-state channel & 22 & 11 & 16 & 12 & 1 & 62 \\
\bottomrule
\end{tabular}
\end{table}

\paragraph{Lever tasks on ML45.} On the two lever tasks the requested effect is about as large as the placement
variation it is read through: the object moves 6.3\,cm in a door-lock demonstration and 8.2 in a door-unlock
one, while the lever's placement varies by 6.3 and 6.2\,cm across episodes (3.3 to 8.2 times the spread on the
other three classes). On door-unlock the hand approaches the lever to 6.5\,cm with the program and to 5.0 for
the reference. On door-lock both reach the lever equally closely (4.5 and 4.7\,cm), and how far the lever is
turned decides the episode: 4.8\,cm in successful episodes, 3.3 and 3.9 in the others. The time pointer gives
11 of 30 on door-lock and the stage pointer 5.

\section{RoboCasa}
\label{app:robocasa}
RoboCasa \citep{robocasa} varies what LIBERO holds fixed. Its 120 procedurally generated kitchens change
layout and style every episode, the robot is a mobile manipulator with a 12-dimensional action, the assets
differ, and the benchmark defines 25 atomic tasks. We convert its 1{,}382 released human demonstrations with
the state-replay procedure used for LIBERO. Three tasks are held out, one per skill family, chosen before any
result was computed: PnPCounterToMicrowave, OpenSingleDoor (whose reverse is in training) and TurnOnStove (a
rotation of a fixed object, like the LIBERO knob).

\paragraph{Adapting the program.} A kitchen contains over a hundred fixtures, so the program keeps only
the movable objects and the joints of the fixtures the task names (RoboCasa's \texttt{fixture\_refs}, or the
task's fixture objects where that annotation is missing). RoboCasa places its world frame at a kitchen
corner; following the LIBERO convention, we therefore express programs, entity poses and end-effector
channels in the mobile base's frame at the episode's first state. A few keypoints lie metres away, so
position channels are normalised by pooled 1st and 99th percentiles. Each held-out episode is rebuilt from
its demonstration's scene XML, first state and cameras, and two checks confirm the rebuild: live keypoints
match the program's first frame to within $10^{-4}$\,mm, and replaying the recorded actions through the
evaluator reaches RoboCasa's success check. Textures were not recorded, so each episode draws its own from
the same library, with the same seed across variants. Step limits are RoboCasa's horizons (600, 500, 500);
everything else follows the LIBERO protocol.

\paragraph{Recipe and results.} The variant uses the \method{} recipe, with the contact channel and the
paired corpus, trained on the remaining 22 tasks; each entity's box is built from its rigid subtree, since
RoboCasa attaches a cabinet door's panel to child bodies. The reference is the same recipe trained on the
observation alone. The effect program completes 21, 24 and 24 of 30 episodes on the stove, the door and the
microwave, 69 of 90 against 44 for the reference and 50 for the terminal state alone
(Table~\ref{tab:robocasa}). On the 49 validation episodes of the training tasks, the effect program, the
terminal state alone and the reference complete 32, 27 and 35.

\begin{table}[h]
\centering\footnotesize
\caption{RoboCasa, the three held-out tasks, 30 episodes each, scored by RoboCasa's own success check. The first four rows share the \method{} recipe and the pointer of \S\ref{sec:infer} and were evaluated on one A100; the third row is the same recipe trained on the observation alone. The last row is a released checkpoint trained only on LIBERO, run zero-shot on the same episodes (see the OpenWAM-$\alpha$ paragraph).}
\label{tab:robocasa}
\setlength{\tabcolsep}{4pt}
\begin{tabular}{lrrrr}
\toprule
 & TurnOnStove & OpenSingleDoor & PnPCounterToMicrowave & $\Sigma/90$ \\
\midrule
effect program & \best{21} & \best{24} & \best{24} & \best{69} \\
terminal state only & 18 & 15 & 17 & 50 \\
reference: observation only & 14 & 14 & 16 & 44 \\
\midrule
\begin{tabular}[t]{@{}l@{}}demonstrator's execution as query\\ (inverse dynamics)\end{tabular} & 9 & 7 & 11 & 27 \\
\midrule
\begin{tabular}[t]{@{}l@{}}OpenWAM-$\alpha$, released LIBERO\\ checkpoint, zero-shot (language)\end{tabular} & 2 & 5 & 3 & 10 \\
\bottomrule
\end{tabular}

\end{table}

\paragraph{OpenWAM-$\alpha$.} OpenWAM-$\alpha$ \citep{openwam} pairs a 5B video backbone with an action
diffusion transformer and conditions on language, two images and the end-effector state. Its released LIBERO
checkpoint was fine-tuned on all four LIBERO suites, which contain our held-out LIBERO tasks, so we evaluate
it on RoboCasa instead, which appears neither in its fine-tuning data nor in its listed pretraining data. It
runs zero-shot on the same 30 episodes per task, with each episode's RoboCasa instruction. The end-effector
state is mapped from RoboCasa's Panda root into LIBERO-Goal's pose (states agree to 0.00\,mm, and nine
held-out demonstrations replay successfully through this mapping), and the mobile base is held still. As an
installation check, it completes 10 of 10 episodes of two LIBERO-Goal tasks. On RoboCasa it completes 2, 5 and 3 of 30 on the stove, the door and the microwave. This row is therefore not a comparison on shared data: the checkpoint has
never seen RoboCasa, and the mapped end-effector height falls outside its state range on 73\% of
demonstration steps.

\section{Programs from real human video}
\label{app:real}
\begin{figure}[h]
\centering
\includegraphics[width=\textwidth]{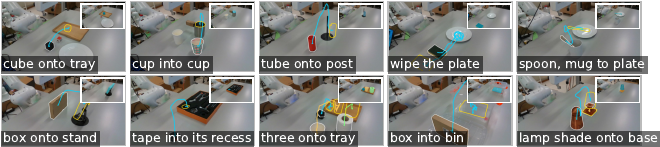}
\caption{\textbf{Programs compiled from ten further real human demonstrations.} Each panel draws the
program on the first frame of the demonstration (white: entities at the start; one colour per entity for its
route; yellow: its terminal region). The inset shows the last frame of the same demonstration.}
\label{fig:real_gallery}
\end{figure}

The recordings of Figures~\ref{fig:real_pair} and~\ref{fig:real_gallery} are RGB videos at 15 to 30\,fps from one fixed camera at the lab table, each showing a person performing one table-top task. We show fourteen such tasks, each moving at least one entity that stays visible through the hand's occlusion. The only human input is one click per
entity on the first frame; SAM~2.1 (Hiera-L) then propagates the mask forward through the video, and a frame
with an empty mask adds no point to the entity's route. Being RGB only, these programs are drawn in image coordinates; the recordings of the four robot tasks of Table~\ref{tab:real} are RGB-D from a calibrated camera, which places each program in the robot's base frame.

A single click carries each mask through the whole video: the masks follow every entity through the hand's occlusion, apart from one frame for the cube of the three-object task and three and eight frames for two of the four balls.

\paragraph{Real-robot setup.} The robot is a Franka Research~3 with its parallel gripper, controlled at 26\,Hz in operational space. An Intel RealSense D435 RGB-D camera, calibrated to the robot's base frame (4\,mm reprojection error of the gripper), records both the human demonstrations and the robot's live observation.
The model is the simulation checkpoint of \S\ref{sec:main} fine-tuned on 156 real-world demonstrations of other tasks, covering 20 object--container and 10 sequential-manipulation combinations, none of them one of the four evaluation tasks; the goal-image variant shares its backbone and training data and is trained with the demonstration's last frame as its condition in place of the program. Each task is demonstrated once by a person; the robot never receives a robot
demonstration of it. A trial succeeds when the manipulated object rests inside the designated target region after the gripper opens; under an edited terminal state, when the cube's final centre lies within 5\,cm of the shifted target, the tolerance of the reversed-program test in Appendix~\ref{app:editing}; both criteria were fixed before the first trial. Every condition of Table~\ref{tab:real} runs 10 trials from object placements within 8\,cm of the demonstration's. On cube to plate the five conditions (goal image, the program as demonstrated, the two shifted terminal states and the push) run in rotation, and the as-demonstrated trials are the \method{} entries of the left block; on the other three tasks the two methods alternate. This gives 110 trials in all. Terminal edits shift
$\Gstar$ by 10\,cm to the left and to the right of the demonstrated placement; the push moves the cube 5\,cm
by hand at the first re-plan after the gripper closes.

\section{RLBench}
\label{app:rlbench}
We evaluate on RLBench \citep{rlbench} because it is the benchmark on which Instant Policy reports its
results, and we hold out all 24 of its tasks. Compared with LIBERO, it changes four things at once: the
simulator is CoppeliaSim rather than MuJoCo, the native action is the next absolute end-effector pose reached
through IK, demonstrations come from scripted planners, and the training set can only contain tasks for which
the planners produce demonstrations.

\paragraph{Training data and filters.} Each candidate training task gets 30 planner attempts within 45
minutes, and a task enters training once the planners reach 20 successes (49 tasks did). The program contains
every dynamic or jointed Shape under the task's base, merged by rigid body and described by the nine box
keypoints in the robot base frame; in scenes with more than 10 entities we keep the ten nearest the first
grasped object. Two checks then remove tasks whose data would be unreliable: a training task is dropped if
replaying it through IK succeeds less than half the time, or if its reset first frame lies 1\,mm or more from
the program. This leaves 39 training tasks, 1{,}136 episodes and 169{,}960 training windows, none of them
from a held-out task.

One recording detail affects two tasks. RLBench derives \texttt{gripper\_open} from the finger opening, so in
CloseJar and OpenJar, where the fingers close on a thick lid, the recorded channel never leaves \emph{open}.
For these two tasks the replay reference and Instant Policy therefore take the gripper state from the
demonstration's grasp record. Our variants read the effect program instead, so they are unaffected.

\paragraph{Evaluation.} Each held-out episode is reset to the stored state of its own demonstration, and
that demonstration serves as the program. We chose the time pointer on validation episodes of the training
tasks (21 of 45 against 13). Episodes use $K_{\text{exec}}{=}4$, a deterministic seed, and a budget of 500
steps or twice the demonstration length. When the IK cannot reach a requested pose, that step is skipped and
the model re-plans at the next boundary; the same rule applies to our variants and to Instant Policy.

\begin{table}[h]
\centering\footnotesize
\caption{RLBench: successes out of 30 on each of the 24 held-out tasks (720 in all), with one demonstration
per episode. The same recipe trains two action heads, one predicting the 7-D end-effector increment
(amendment G, the configuration reported in \S\ref{sec:beyond}) and one predicting the next absolute
end-effector pose. Program: the effect program. Relative: the effect written as displacements from each
entity's anchor. Reference: the same recipe trained on the observation alone. IP: Instant Policy with one
and two demonstrations. Replay: the demonstration's own absolute-pose actions executed through the same IK
(replaying the increments instead completes 616), with CloseJar and OpenJar using the gripper state from the
grasp record. Every evaluation uses the clock pointer, $K_{\text{exec}}{=}4$ and one sampling seed, and counts
an infeasible pose as a skipped step.}
\label{tab:rlbench}
\scriptsize\setlength{\tabcolsep}{2.3pt}
\begin{tabular}{lrrrrrrrrr}
\toprule
 & \multicolumn{3}{c}{increment action} & \multicolumn{3}{c}{absolute pose} & & & \\
\cmidrule(lr){2-4}\cmidrule(lr){5-7}
 & program & relative & reference & program & relative & reference & IP, 1 demo & IP, 2 demos & replay \\
\midrule
LampOn & 29 & 12 & 30 & 0 & 25 & 0 & 30 & 30 & 30 \\
PutRubbishInBin & 27 & 0 & 2 & 8 & 1 & 0 & 30 & 30 & 30 \\
PushButton & 26 & 6 & 28 & 20 & 12 & 12 & 29 & 30 & 30 \\
ToiletSeatDown & 20 & 12 & 14 & 19 & 15 & 11 & 30 & 29 & 30 \\
TakeLidOffSaucepan & 17 & 15 & 10 & 0 & 20 & 27 & 29 & 30 & 30 \\
TakeUmbrellaOutOfUmbrellaStand & 15 & 7 & 16 & 6 & 8 & 12 & 29 & 30 & 29 \\
PressSwitch & 15 & 11 & 2 & 14 & 6 & 6 & 30 & 28 & 30 \\
BasketballInHoop & 4 & 0 & 0 & 0 & 0 & 0 & 30 & 30 & 30 \\
TakeToiletRollOffStand & 3 & 2 & 3 & 13 & 2 & 5 & 30 & 30 & 30 \\
PhoneOnBase & 2 & 11 & 1 & 7 & 23 & 0 & 30 & 30 & 30 \\
SlideBlockToTarget & 1 & 9 & 0 & 0 & 0 & 0 & 30 & 30 & 30 \\
OpenJar & 1 & 2 & 0 & 2 & 2 & 0 & 0 & 0 & 1 \\
CloseMicrowave & 0 & 14 & 0 & 7 & 15 & 19 & 30 & 30 & 30 \\
OpenBox & 0 & 1 & 0 & 0 & 0 & 0 & 30 & 30 & 30 \\
CloseBox & 0 & 9 & 0 & 1 & 0 & 0 & 26 & 30 & 30 \\
PutUmbrellaInUmbrellaStand & 0 & 0 & 0 & 0 & 0 & 0 & 16 & 21 & 4 \\
TakePlateOffColoredDishRack & 0 & 1 & 0 & 1 & 2 & 0 & 30 & 29 & 30 \\
CloseLaptopLid & 0 & 0 & 0 & 0 & 0 & 0 & 29 & 30 & 30 \\
BeatTheBuzz & 0 & 6 & 1 & 0 & 2 & 0 & 3 & 28 & 30 \\
ToiletSeatUp & 0 & 2 & 1 & 0 & 0 & 0 & 29 & 30 & 30 \\
OpenMicrowave & 0 & 0 & 0 & 0 & 0 & 0 & 0 & 15 & 30 \\
MeatOnGrill & 0 & 1 & 0 & 0 & 0 & 1 & 26 & 26 & 25 \\
MeatOffGrill & 0 & 2 & 0 & 0 & 1 & 0 & 20 & 25 & 30 \\
CloseJar & 0 & 0 & 0 & 11 & 0 & 0 & 9 & 9 & 7 \\
\midrule
total of 720 & \best{160} & 123 & 108 & 109 & 134 & 93 & 575 & 630 & 636 \\
\bottomrule
\end{tabular}

\end{table}

\paragraph{Two action heads.} Our first variants predicted the next absolute end-effector pose, which is
RLBench's native action but not the one our model uses on LIBERO. Amendment G therefore retrains the three
variants with the LIBERO action, a 7-D end-effector increment (position scaled by 0.025\,m and rotation by
0.10\,rad, gripper last), converted from the same demonstrations. The increment is added to the commanded
setpoint, not to the measured pose: the controller lags the command by 1.5 to 4\,mm per step, and
integrating the measured pose would accumulate that lag (amendment G-2). The setpoint is reset to the
measured pose at the start of an episode, after an IK rejection, and whenever the two differ by more than
3\,cm or $15^\circ$. As an upper reference for this execution path, replaying the held-out demonstrations
through it completes 616 of 720, against 636 for absolute-pose replay.

\paragraph{Results.} Table~\ref{tab:rlbench} gives the per-task counts. With the increment head, the effect
program completes 160 of the 720 held-out episodes, while the same recipe without a program completes 108.
The margin of 52 exceeds the 48 fixed in advance. The relative program lies in between, at 123. Four tasks
account for most of the margin: PutRubbishInBin (27 against 2), PressSwitch (15 against 2), ToiletSeatDown
(20 against 14) and TakeLidOffSaucepan (17 against 10).

\begin{figure}[h]
\centering
\includegraphics[width=0.9\textwidth]{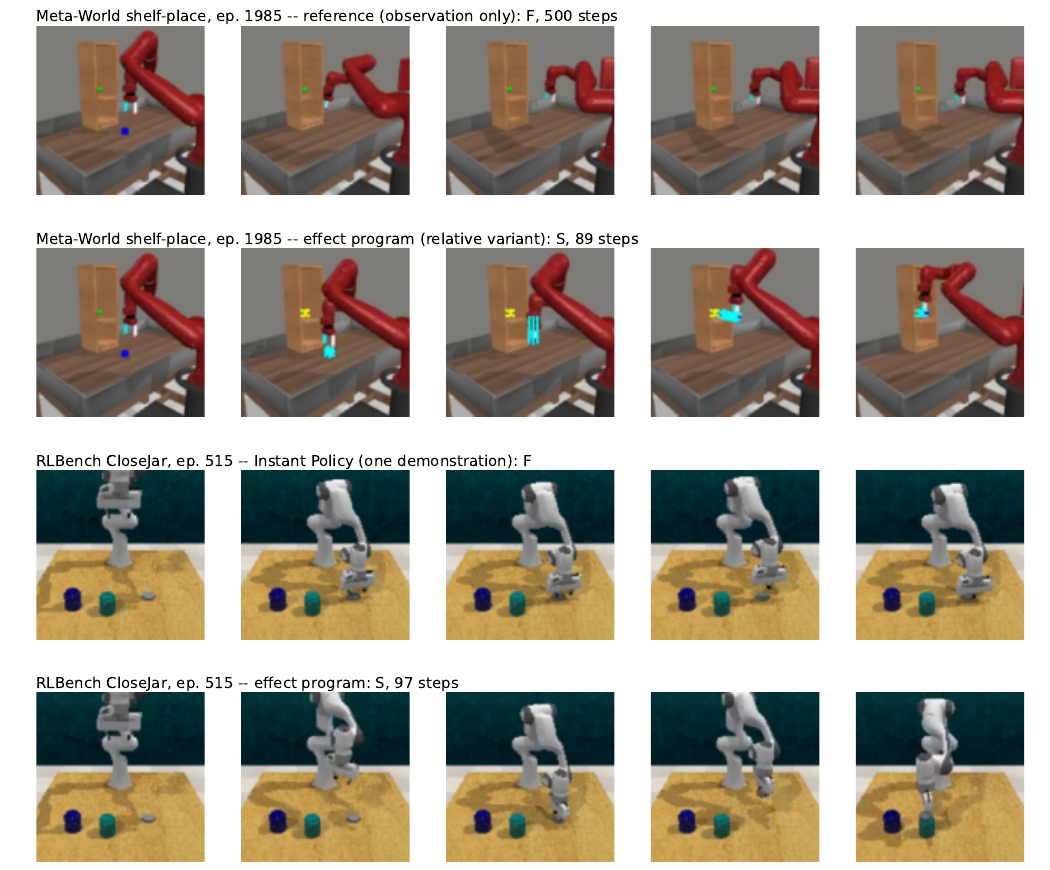}
\caption{For Meta-World and RLBench, the first episode in the catalogue on which
the baseline fails and the effect program succeeds. Each row shows five frames of one rollout, first to last,
taken from that method's own recording, with the outcome stored in its own record. Meta-World: the reference, given
no task specification, descends to the block and retreats; the relative variant lifts the block onto the
shelf. RLBench: Instant Policy, given segmented point clouds and one demonstration, hovers over the lid; the
effect program places the lid on the jar.}
\label{fig:showcase}

\end{figure}

The absolute-pose head orders the variants differently. Here the relative program leads with 134, followed by
the program with 109 and the reference with 93. On the two lid tasks the absolute-pose program does better
than Instant Policy (CloseJar 11 against 9, OpenJar 2 against 0; Figure~\ref{fig:showcase}). Over all tasks,
Instant Policy completes 575 with one demonstration and 630 with two.

\paragraph{Execution on RLBench.} Two measurements characterise the episodes: those in which no entity moves,
and requested poses the IK cannot reach. Episodes with no entity moving by a centimetre go from 398 of 720 with
the absolute-pose program to 338 with the increment program, and the share of executed steps whose requested
pose the IK cannot reach is 0.283 for the increment program against 0.303. Unreachable poses concentrate in a
subset of tasks: in the absolute-pose variants a task with no successes requests 6{,}800 to 9{,}400 of them,
while LampOn, TakeLidOffSaucepan and PutRubbishInBin request none. The program's first frame lies a median of
1\,mm or less from the scene on 22 tasks.

\paragraph{A fallback for rejected poses (amendment H).} Since rejected poses are this common, amendment H
changes what the executor does with them. Previously the arm simply held in place. With the fallback, the
executor first retries with the step from the setpoint scaled by $1/2$, $1/4$ and $1/8$, then keeps the
position and holds the orientation at the measured one, and only then holds in place. Weights and protocol
are unchanged, and the three increment variants run the same 720 episodes.

The fallback lowers the share of executed steps that end in a hold from 0.283 to 0.185 for the program variant and from 0.380 to 0.236 for the reference (Table~\ref{tab:rlbench_fb}). The reference rises from 108 to 121 successes, gaining on
ToiletSeatDown, TakeLidOffSaucepan and CloseMicrowave, whereas the program rises from 160 to 163, with its
gain on PressSwitch (15 to 19). The margin becomes 42; \S\ref{sec:beyond} reports the configuration without the fallback.

\begin{table}[h]
\centering\small
\caption{RLBench increment variants without and with the fallback for rejected poses (amendment H), on the
24 held-out tasks with 720 episodes per variant. Hold: share of executed steps at which no pose was reachable
and the arm held in place. Still: episodes in which no entity moved by 1\,cm. Per-task counts will be
released with the code.}
\label{tab:rlbench_fb}
\begin{tabular}{lrrrrrr}
\toprule
 & \multicolumn{2}{c}{successes} & \multicolumn{2}{c}{hold} & \multicolumn{2}{c}{still} \\
\cmidrule(lr){2-3}\cmidrule(lr){4-5}\cmidrule(lr){6-7}
variant & without & with & without & with & without & with \\
\midrule
program & 160 & 163 & 0.283 & 0.185 & 338 & 332 \\
relative & 123 & 134 & 0.596 & 0.518 & 387 & 383 \\
reference & 108 & 121 & 0.380 & 0.236 & 316 & 300 \\
\midrule
program $-$ reference & 52 & 42 & & & & \\
\bottomrule
\end{tabular}

\end{table}

\section{Reproduction on a second GPU generation}
\label{app:v100}
This section repeats key results on a second GPU generation to check that they do not depend on the hardware. Its evaluations were run before the engagement alignment of \S\ref{sec:infer} was added and compare the two machines under that earlier pointer. The two
GPU generations, Volta and Ampere, disagree on borderline episodes: of three we checked by hand, two are
bit-identical and one succeeds on the V100 but fails on the 3090. We therefore compare numbers from the V100
machine only with variants re-run on that machine.

Under this rule, the findings of \S\ref{sec:edit} and \S\ref{sec:analysis} reproduce
(Table~\ref{tab:v100}). The first-seed pair still separates on following (0.78 against 0.33), and the
weight-encoded model drifts with the same sign-independent offset. We also re-ran the two headline queries at
LIBERO's ceiling of 50 episodes per task. The effect program completes 43, 19 and 1 of 50 and the terminal
state 50, 35 and 0, the same ordering as at 30 episodes. Finally, a second training of the
weight-encoded model differs from the first by 22 on the bowl sum over the four arms.

\begin{table}[h]
\centering\small
\caption{Re-evaluation on three V100s. $\Sigma$bowl and
$\Sigma$bottle are the sums over the four arms, as in Table~\ref{tab:arms}; following is the mean over the
four in-distribution commanded rotations. The last column repeats the corresponding 3090 number. Following is undefined for the weight-encoded replicate, whose realised
rotation is negative for every commanded rotation ($-60$, $-70$, $-49$ and $-77^\circ$ for $+20$, $-20$,
$+45$ and $-45^\circ$).}
\label{tab:v100}
\begin{tabular}{lrrrrrrl}
\toprule
run (V100) & Panda & IIWA & Kinova3 & UR5e & $\Sigma$bowl & following & 3090 \\
\midrule
\method{}            & 26/10 & 20/18 & 20/6 & 16/6 & 82 & 0.78 & 81, \ 0.73 \\
W8                   & 28/20 & 18/15 & 22/11 & 2/1 & 70 & 0.33 & 68, \ 0.38 \\
v14 (weights) replicate & 23/11 & 18/10 & 15/6 & 8/0 & 64 & n/a & 86 \\
\bottomrule
\end{tabular}

\end{table}

\section{Qualitative results and videos}
\label{app:gallery}

\begin{figure}[h]
\centering
\includegraphics[width=\textwidth]{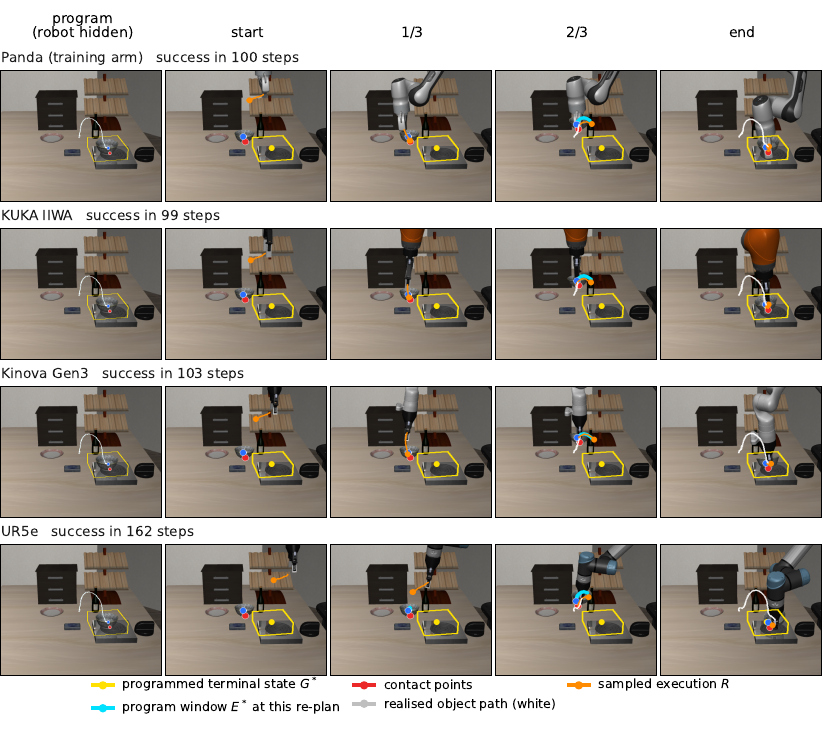}
\caption{One program executed on four arms without retraining (episode 227, side view). The episodes take
100, 99, 103 and 162 steps, as in the evaluation records.}
\label{fig:arms}

\end{figure}

\begin{figure}[h]
\centering
\includegraphics[width=\textwidth]{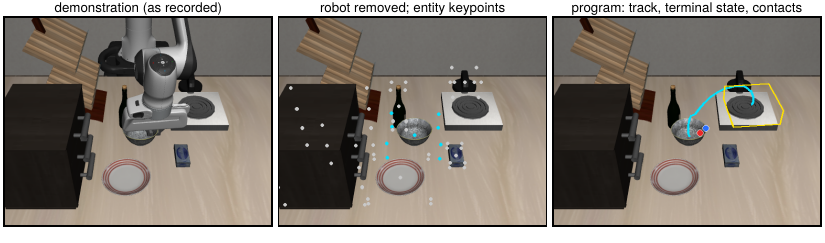}
\caption{From a demonstration to a program (bowl task, episode 201). Left to right: the demonstration as
recorded; the same frame with the robot removed, showing the nine keypoints of every entity; and the program
built from them, consisting of the object's track, its terminal state and the two contact points.}
\label{fig:program}

\end{figure}

\begin{figure}[h]
\centering
\includegraphics[width=\textwidth]{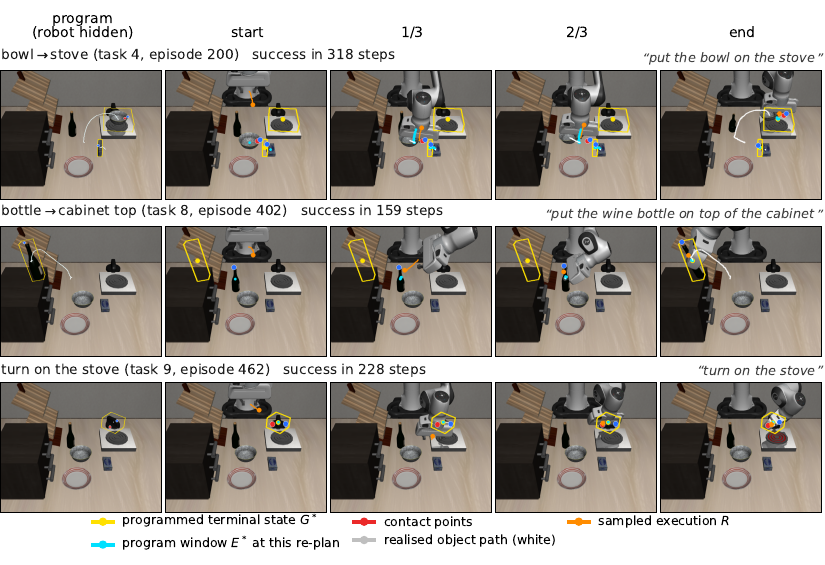}
\caption{Figure~\ref{fig:heldout} with five moments per episode: the program, then the execution at its
start, one third, two thirds and end. Each row shows the outcome stored in its evaluation record.}
\label{fig:heldout_full}

\end{figure}

\begin{figure}[h]
\centering
\includegraphics[width=\textwidth]{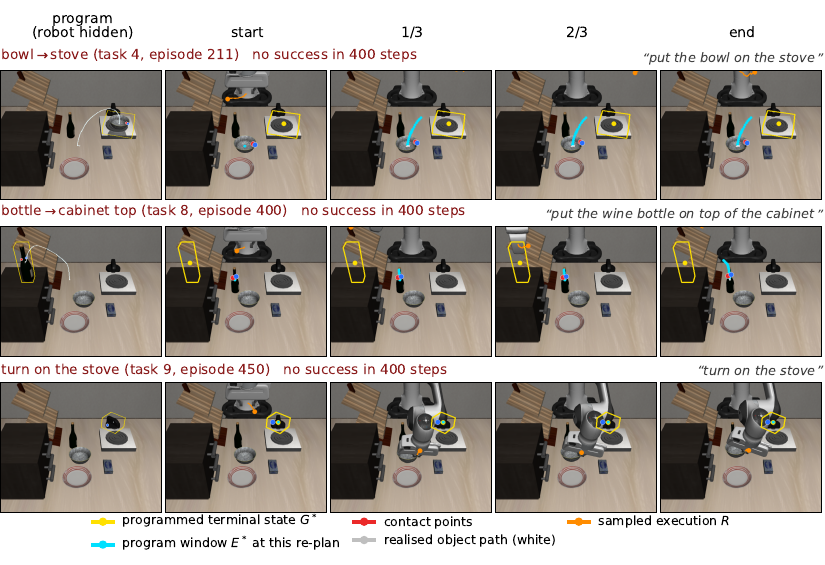}
\caption{The lowest-numbered failure of each held-out task in the same record (episodes 211, 400 and 450), drawn as in Figure~\ref{fig:heldout_full}. Each runs to the 400-step limit.}
\label{fig:heldout_fail}

\end{figure}

\begin{figure}[h]
\centering
\includegraphics[width=0.97\textwidth]{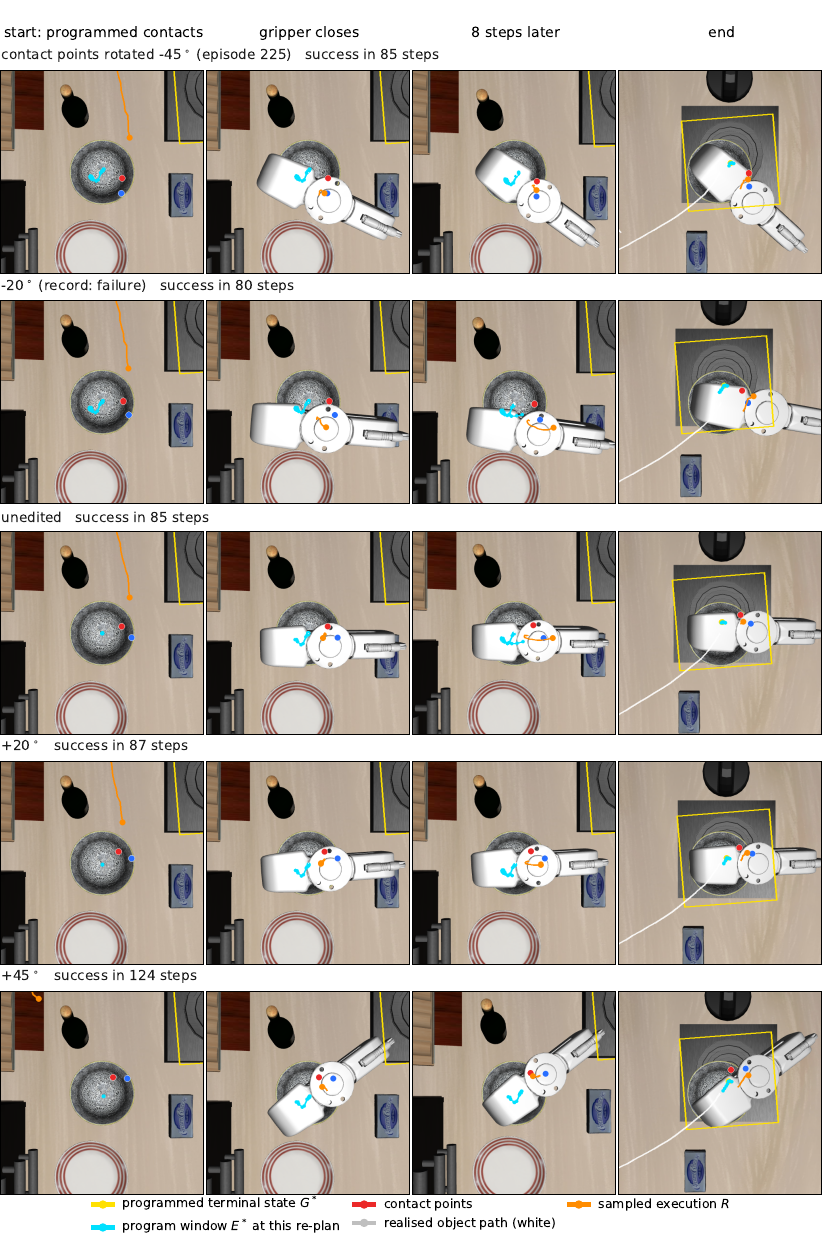}
\caption{The contact points rotated to five commanded angles on episode 225, seen from above: the
programmed contacts, the frame at which the gripper closes, eight steps later, and the end. In this re-run the $-20^\circ$ rotation succeeds, while its record fails; the other four rotations reproduce their records.}
\label{fig:rungs}

\end{figure}

\begin{figure}[h]
\centering
\includegraphics[width=\textwidth]{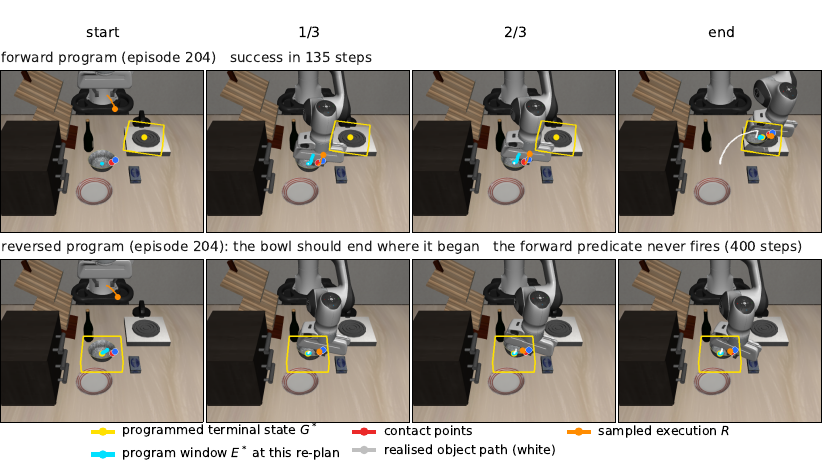}
\caption{The program and the same program reversed, on episode 204, both started from the demonstration's
initial state. The forward program puts the bowl on the stove; under the reversed program the arm reaches the bowl and leaves it in place.}
\label{fig:reverse}

\end{figure}

\begin{figure}[h]
\centering
\includegraphics[width=0.66\textwidth]{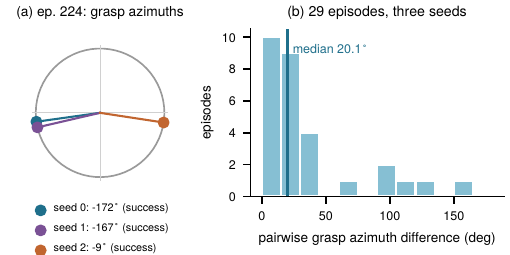}\\[2pt]
\includegraphics[width=\textwidth]{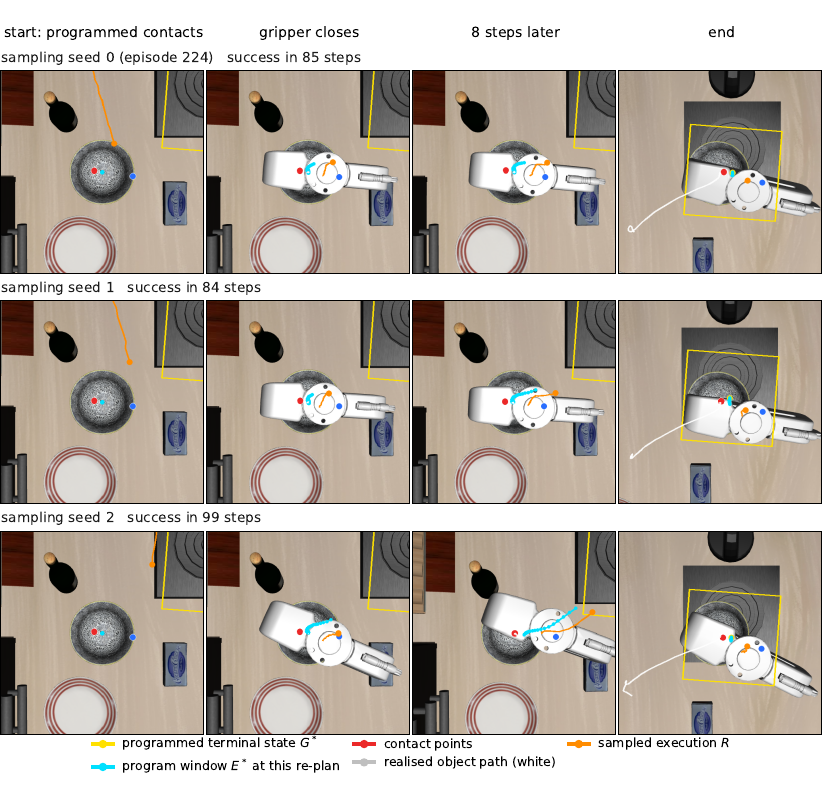}
\caption{Three sampling seeds on one program. Top: (a) the grasp azimuth each seed realises on episode
224, where all three succeed and seed 2 takes the bowl from the other side; (b) the median of the three
pairwise azimuth differences for each of the 29 episodes with a grasp under all three seeds. Bottom: episode
224 under each seed, seen from above.}
\label{fig:seeds}

\end{figure}

\begin{figure}[h]
\centering
\includegraphics[width=0.66\textwidth]{figs/fig_q5_plot.pdf}\\[2pt]
\includegraphics[width=\textwidth]{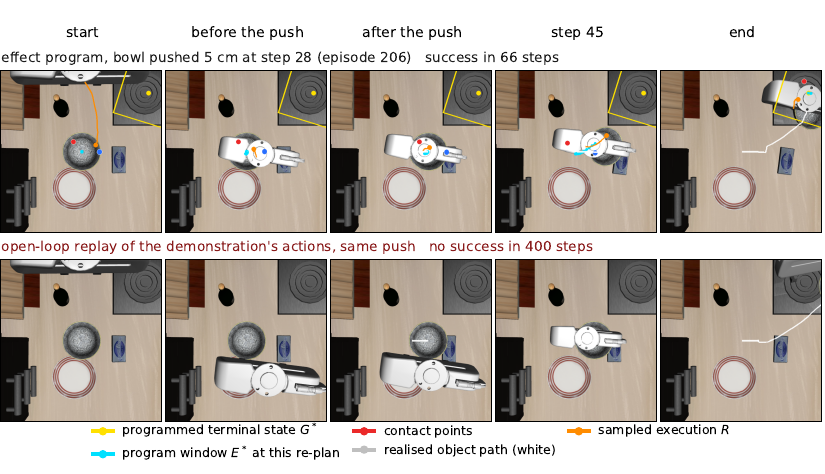}
\caption{A 5\,cm push at step 28 of episode 206. Top: the bowl's distance to the programmed end over time.
Despite the push, the effect program succeeds in 66 steps (95 undisturbed). Open-loop replay of the
demonstration's actions succeeds in 93 steps when undisturbed, but after the push it ends 8.0\,cm from the
programmed end after 400. Bottom: the two pushed episodes, seen from above. We re-ran these episodes for this figure to record the trajectory; they reproduce the outcomes and step counts of the records behind Table~\ref{tab:disturb}.}
\label{fig:push}

\end{figure}

For each held-out task, the representative episode is the lowest-numbered episode that the headline model (seed 0) completes in the record behind Table~\ref{tab:main}, shown next to its lowest-numbered failure (Figures~\ref{fig:heldout_full} and~\ref{fig:heldout_fail}). The commanded rotations are shown on episode 225, on which the unedited program succeeds (Figure~\ref{fig:rungs}); the four arms on episode 227, one of the 13 episodes that all four arms complete (Figure~\ref{fig:arms}); the reversed program on episode 204, on which the forward program succeeds (Figure~\ref{fig:reverse}); the three sampling seeds on episode 224 (Figure~\ref{fig:seeds}) and the push on episode 206 (Figure~\ref{fig:push}). Figure~\ref{fig:program} shows the compilation of a program from a demonstration.

To render these episodes, we re-ran each one with the evaluator unchanged and a per-step trace added, then
replayed it offline from its recorded actions at $512^2$. The replayed keypoints equal the traced ones at
every step, so the images show exactly what the evaluator executed. Every rendered episode reproduces its recorded outcome except the $-20^\circ$ rotation of Figure~\ref{fig:rungs}, which succeeds in the re-run and fails in the record; the figure labels it as such. All overlays use one colour convention: yellow is
$\Gstar$, cyan the window of $\Estar$ clamped at that re-plan, red and blue the contact points, orange the
sampled execution $R$, and white the object's realised path.

\paragraph{Videos.} The supplementary material contains one real-robot video. It shows the cube-to-plate and
cube-to-box programs of Table~\ref{tab:real}, the terminal-state edits, and further tasks of
Figure~\ref{fig:real_gallery} and beyond, each programmed by one human video and executed closed loop; beyond
the counts of Table~\ref{tab:real}, these runs are qualitative. The recordings behind every figure, and a
catalogue of held-out Meta-World and RLBench episodes on which a baseline fails and the effect program
succeeds, will be released with the code upon acceptance. Each clip was checked against its own evaluation record. Where an episode had no recording, we re-ran it
alone with the record's flags and kept it only if outcome and step count matched; all seven such re-runs
did. The catalogue holds 132 Meta-World pairs against the reference and 4 RLBench pairs against Instant Policy.
The released gallery videos show the four arms on one episode, the $3\times3$ grid of contact rotation and
terminal shift, the five commanded rotations, three sampling seeds, the push, and the reversed program.

\begin{figure}[h]
\centering
\includegraphics[width=\textwidth]{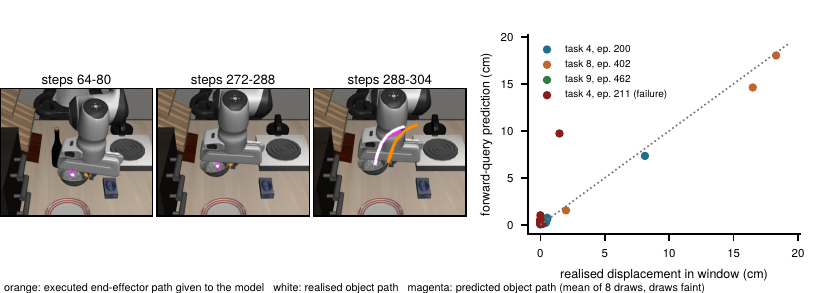}
\caption{The forward query. Left: three consecutive windows of episode 200, showing the executed
end-effector path given to the model (orange), the object's realised path (white) and its predicted path
(magenta: the mean of eight draws, with the individual draws faint). Right: predicted against realised
displacement for each window of four episodes.}
\label{fig:forward}

\end{figure}
\begin{figure}[h]
\centering
\includegraphics[width=\textwidth]{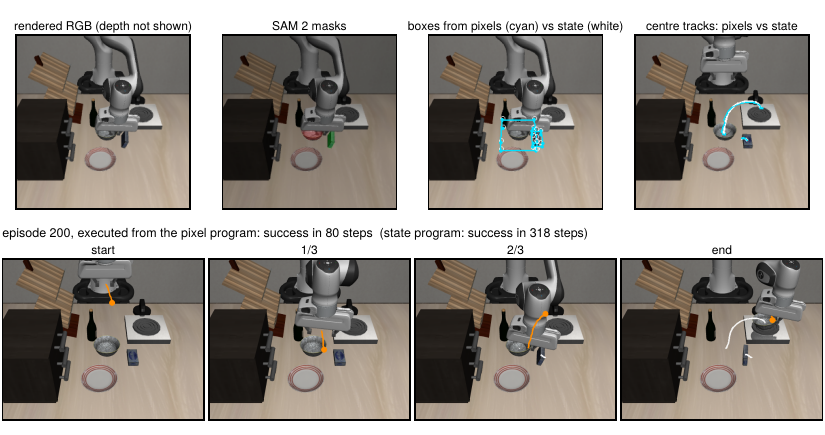}
\caption{A program from pixels (bowl task, episode 200, the first episode of the paired comparison of
\S\ref{sec:edit}). Top: the rendered frame, the SAM~2 masks, the oriented boxes fitted from pixels against
the simulator's, and the object centre tracked from pixels against state (median error 0.61\,cm for the
bowl, at most 1.51). Bottom: the pixel program executed. This episode succeeds in 80 steps from pixels and in 318 from state; over the 30 paired episodes, 26 succeed from pixels and 27 from state.}
\label{fig:pixels}

\end{figure}
\begin{figure}[h]
\centering
\includegraphics[width=\textwidth]{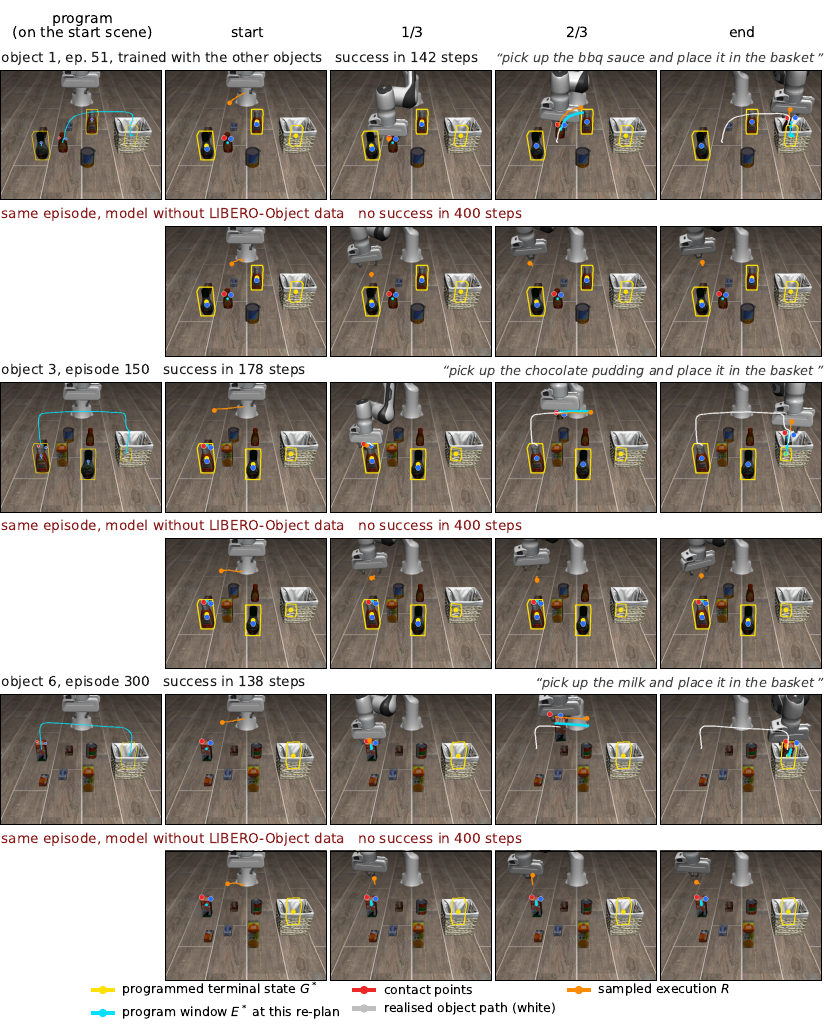}
\caption{LIBERO-Object, the three held-out objects (\S\ref{sec:beyond}): the lowest-numbered success of
each in the record of Table~\ref{tab:object}, and below it the same episode under the reference, which is
trained without LIBERO-Object data and fails on all three. The program is drawn on the episode's start
scene. These reruns reproduce every outcome of the table's records.}
\label{fig:object}

\end{figure}

\paragraph{The forward query.} The forward query asks the model how the object will move under a given
execution. We pose it at every 16-step window of four episodes (the three successes of
Figure~\ref{fig:heldout} and the failure of episode 211): the model receives the live scene and the
end-effector execution actually carried out, and predicts the object's motion (Figure~\ref{fig:forward}).
Across 67 windows, predicted and realised displacement magnitudes correlate at 0.94. The object moves in five windows, by a median of 8.1\,cm, and there the median error is 1.5\,cm; in the windows where it stays, the median error is 0.30\,cm. In episode 211 the bowl is pushed by at most 1.7\,cm and never lifted.

\end{document}